\documentclass{article}

\PassOptionsToPackage{}{natbib}

    \usepackage[final]{neurips_2020}

\usepackage[utf8]{inputenc} 
\usepackage[T1]{fontenc}    

\usepackage[pagebackref=true,breaklinks=true,colorlinks=true,bookmarks=false,urlcolor=black, allcolors=mydarkblue]{hyperref}
\usepackage{url}            
\usepackage{booktabs}       
\usepackage{amsfonts}       
\usepackage{nicefrac}       
\usepackage{microtype}      
\usepackage{multirow}
\usepackage{adjustbox}
\usepackage{caption}
\usepackage{float}
\usepackage{mathtools}  
\usepackage{caption}
\usepackage{subcaption}  
\usepackage{listings}   
\usepackage{booktabs}   
\usepackage{upgreek}
\usepackage[para]{footmisc}
\usepackage{xspace}

\newcommand{\Ezero}{FP8-E0\xspace}
\newcommand{\Eone}{FP8-E1\xspace}
\newcommand{\Etwo}{FP8-E2\xspace}
\newcommand{\Ethree}{FP8-E3\xspace}
\newcommand{\Efour}{FP8-E4\xspace}
\newcommand{\Efive}{FP8-E5\xspace}

\usepackage{changepage}

\makeatletter
\newcommand{\printfnsymbol}[1]{%
  \textsuperscript{\@fnsymbol{#1}}%
}
\makeatother

\makeindex

\usepackage{version}

\title{FP8 versus INT8 for efficient deep learning inference}

\author{%
  Mart van Baalen \\ 
  Qualcomm AI Research\thanks{Qualcomm AI Research is an initiative of Qualcomm Technologies, Inc.}\\
  \texttt{mart@qti.qualcomm.com}
  \And
  Andrey Kuzmin \\ 
  Qualcomm AI Research\footnotemark[1]\\
  \texttt{akuzmin@qti.qualcomm.com} \AND
  Suparna S Nair \\ 
  Qualcomm AI Research\footnotemark[1]\\
  \texttt{suparnas@qti.qualcomm.com}
  \And
  Yuwei Ren \\ 
  Qualcomm AI Research\footnotemark[1]\\
  \texttt{ren@qti.qualcomm.com}
  \AND
  Eric Mahurin \\ 
  Qualcomm AI Research\footnotemark[1]\\
  \texttt{mart@qti.qualcomm.com}
  \And
  Chirag Patel \\ 
  Qualcomm AI Research\footnotemark[1]\\
  \texttt{cpatel@qti.qualcomm.com}
    \AND
  Sundar Subramanian \\ 
  Qualcomm AI Research\footnotemark[1]\\
  \texttt{sundars@qti.qualcomm.com}
    \And
  Sanghyuk Lee \\ 
  Qualcomm AI Research\footnotemark[1]\\
  \texttt{sankee@qti.qualcomm.com}
    \And
  Markus Nagel \\ 
  Qualcomm AI Research\footnotemark[1]\\
  \texttt{mnagel@qti.qualcomm.com}
    \And
  Joseph Soriaga \\ 
  Qualcomm AI Research\footnotemark[1]\\
  \texttt{jsoriaga@qti.qualcomm.com}
    \And
  Tijmen Blankevoort \\ 
  Qualcomm AI Research\footnotemark[1]\\
  \texttt{tijmen@qti.qualcomm.com}
 }

\author{%
  Mart van Baalen \quad Andrey Kuzmin \quad Suparna S Nair \quad Yuwei Ren \quad Eric Mahurin \\ \textbf{Chirag Patel} \quad \textbf{Sundar Subramanian} \quad \textbf{Sanghyuk Lee} \quad \textbf{Markus Nagel} \quad 
  \textbf{Joseph Soriaga} \\
  \textbf{Tijmen Blankevoort} \\ 
  Qualcomm AI Research\thanks{Qualcomm AI Research is an initiative of Qualcomm Technologies, Inc.}\\
  \texttt{\{mart, akuzmin, suparnas, ren, emahurin, cpatel, sundars, } \\
  \texttt{sanlee, markusn, jsoriaga, tijmen\}@qti.qualcomm.com}
 }

\begin{document}

\maketitle

\section{Introduction} 

Recently, the idea of using FP8 as a number format for neural network training has been floating around the deep learning world. Nvidia has announced the FP8 format for their transformer engine software for the new Hopper architecture GPUs, and an IEEE consortium is investigating the standardization of the FP8 format for training deep learning networks in the cloud. Given that most training is currently conducted with entire networks in FP32, or sometimes FP16 with mixed-precision, the step to having some parts of a network run in FP8 with 8-bit weights is an appealing potential speed-up for the generally costly and time-intensive training procedures in deep learning.

A natural question arises regarding what this development means for efficient inference on edge devices. In the efficient inference device world, workloads are frequently executed in INT8. Sometimes going even as low as INT4 when efficiency calls for it. Porting FP32 or FP16-trained models to the INT format is called quantization. This quantization conversion step is not always straightforward and sometimes requires a bit of effort (\cite{whitepaper}), but in many scenarios, the work is worth it to get to networks that are 2 to 8 times more efficient than their FP32 counterparts. Due to the time it sometimes takes to quantize networks, it is certainly an alluring idea that one could train entire networks in FP8 and then deploy them in the same format and get similar efficiency. One would hypothetically have the same benefits as the INT format without the hassle of undertaking quantization after training.

This whitepaper explains why this scenario will likely fail to come to fruition. First, we show that in the compute of dedicated hardware, the FP8 format is at least 50\% less efficient in terms of area and energy usage than INT8. This means that FP8 will have to be significantly more accurate than INT8 to be worthwhile from a hardware-efficiency perspective. We also compare the performance in terms of network accuracy for the generally proposed FP8 formats with 4 and 5 exponent bits with INT8. Based on our recent paper on the FP8 format (\cite{fp8paper}), we theoretically show the difference between the INT8 and FP8 formats for neural networks and present a plethora of post-training quantization and quantization-aware-training results to show how this theory translates to practice. Based on our research and a read of the research field, we conclude that although the proposed FP8 format is potentially a good match for gradients during training (although comparative evidence to other formats is sparse), the results for inference do not warrant a dedicated implementation of FP8 in favor of INT8. 

Later, we compare our findings to other works in the literature. We show that our results are mostly consistent with previous findings but that important comparisons between the formats have thus far been lacking. Finally, we discuss what happens when FP8-trained networks are converted to INT8 and conclude with a brief discussion on the most efficient way for on-device deployment and an extensive suite of INT8 results for many models.

\section{Preliminaries} \label{sec:preliminaries} 
Comparing the floating point with the integer format, one difference is the possibility of having a number of exponent bits. Integer numbers $z$ can be represented as

\begin{align*}
    z &= s \cdot \hat{z} + b \\
    \hat{z} &= z_1 \ldots z_8
\end{align*}

$s$ is the scale parameter, and $b$ is an optional bias, the $z_i$ values are individual bits. The floating-point numbers are similar but allow for a different bit allocation for exponent bits versus mantissa bits.

\begin{align*}
    z &= \left(1+\frac{\hat{m}}{2^M}\right) \cdot 2^{\hat{e}-B}\\
    \hat{m} &= m_1 \ldots m_M \\
    \hat{e} &= e_1 \ldots e_E 
\end{align*}

where $\hat{m}$ is the mantissa, $\hat{e}$ the exponent, $B$ the exponent bias, $M$ the number of mantissa bits and $E$ the number of exponent bits. We assume that our formats include subnormal numbers\footnote{For values where $\hat{e}=0$, we have that $z=\frac{\hat{m}}{2^M} \cdot 2^{1-B}$}, such that exact $0$ can be represented. Furthermore, we assume only positive $0$ is represented and reserve negative $0$ to encode a single special value (e.g., NaN). Lastly, similarly to integer formats, we assume that groups of weights and activations can be scaled arbitrarily and that for weight tensors this scaling can be applied per channel. The actual number format used is slightly more complicated, and a more detailed discussion can be found in Appendix \ref{app:number_description}. 

The commonly proposed formats, which are also implemented in Nvidia’s Hopper architectures, are FP8 with 4 or 5 exponent bits (\cite{hopper_in_depth}). We will denote the number of exponent bits in the paper as FP8-E[X], such that the proposed formats with 4 and 5 exponent bits are referred to as, respectively,  
\Efour and \Efive. We will also investigate what happens with the number formats \Etwo and \Ethree\footnote{Note that the inclusion of subnormal numbers implies that the \Ezero and \Eone formats are identical to INT8.}. This is done to help paint a complete picture and give an understanding of the behavior of these formats. As noted in the FP8 introduction paper from Nvidia (\cite{nvidia_fp8}), the \Efive format is mostly used for the gradients, so in our comparison for efficient inference, we will mostly be focusing on the difference between INT8 and \Efour.  

\section{Hardware Considerations} \label{sec:hardware}

\subsection{FP8 is an ambiguous term} \label{subsec:ambiguous}

There are several choices that can be made in designing dedicated neural network inference hardware. Specifying a number format for a network does not paint the entire picture of what goes on in the hardware. Nor does it give a very clear description of the efficiency of a format. 

\begin{figure}[t]
\centering
\begin{tabular}{c}
\includegraphics[width=10.0cm]{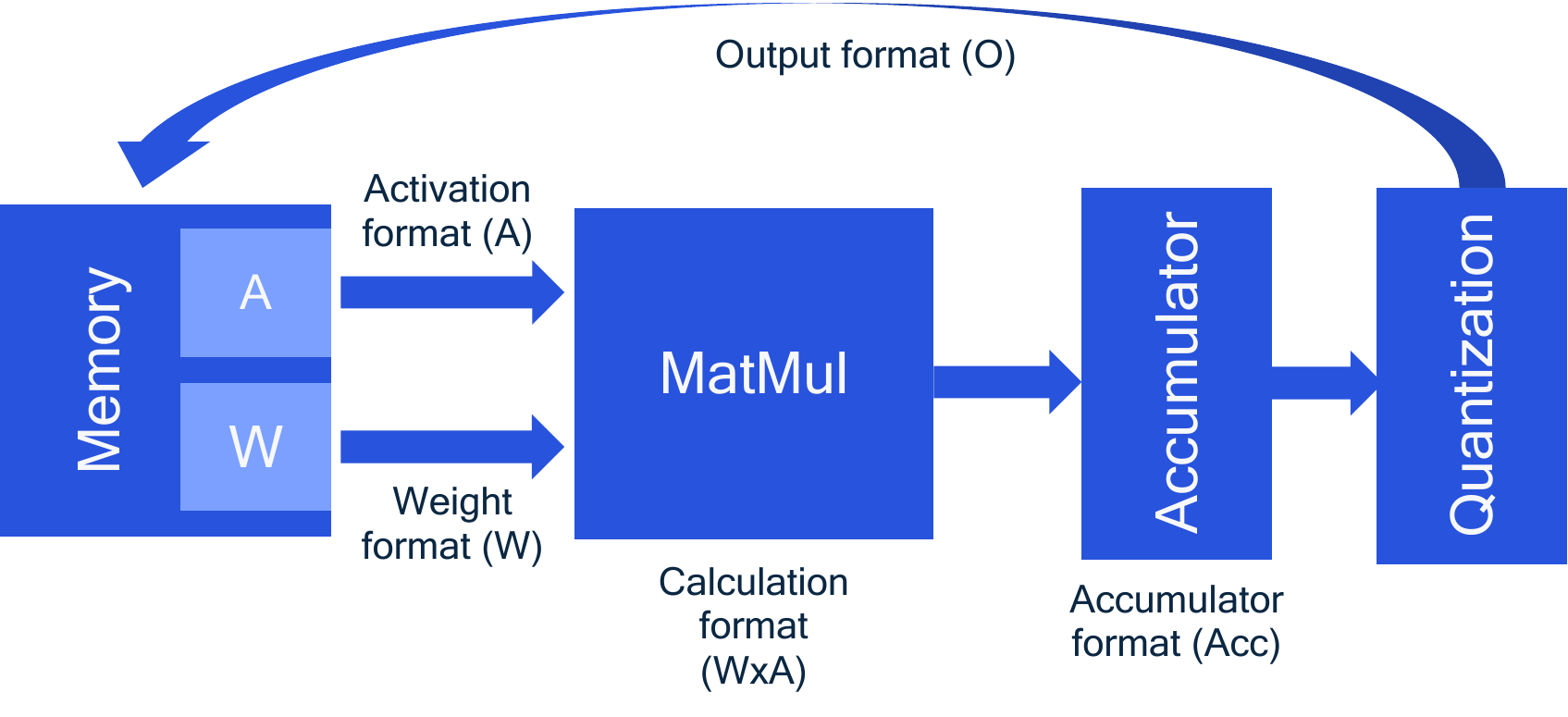}

\end{tabular}
\caption{A schematic overview of a deep learning accelerator. Weights and activations are stored in memory and moved to the MatMul calculation unit. The bit-width matters for both latency and energy consumed for transferring the data. The calculation unit does a matrix multiplication; here, both the bit-width and the format matter for latency and energy consumption. The accumulator stores the intermediate results, for which the format/bit-width can be chosen. Finally, the output format also has a choice, for which the bit-width dictates how many bits are transferred and stored back in memory.}
\label{fig:hardware_sketch}
\end{figure}

We can look at Figure~\ref{fig:hardware_sketch} to understand why. This figure has a schematic overview of a deep-learning accelerator. This coarse picture will apply to most commonly used accelerators. Weights are stored in some form of memory. Inputs and activations are stored in that memory as well. The weights and the activations are loaded into a MatMul calculation unit to calculate dot products for convolutions and matrix multiplication. The results of several calculations are then written to an accumulator with a higher bit-width to sum values with a high precision. Once the computation is done, the values from the accumulator are potentially re-quantized and then written back to memory for further processing. It is possible the calculations are intended to be done in, e.g., 8 bits, but the activation tensors are stored in 16 bits. In this case, the 16-bit tensor will first be converted to an 8-bit tensor before doing the computation. This occurs when a previous layer outputs a 16-bit tensor, and the intermediate operations before the current layer are done in 16-bit. 

\subsection{Different choices lead to different performance} \label{subsec:different_performance}

There is a choice to be made for each of these steps. What are the bit-widths and formats of the weights and the activations, the accumulators, and the values written back to memory? These choices matter; their efficacy depends on the specific architecture and network. In neural network hardware, you are generally either bandwidth or compute-limited. The extent of this depends on the network and hardware itself. In the ideal hardware for inference, you are right on the edge between the two, so no resources are wasted. For example, if you are constrained by memory bandwidth, writing tensors to and from memory in 16 bits instead of 8 bits will hurt inference speed. This frequently happens for networks applied on large images, as the image/activation memory usage becomes very large compared to the computations. Similarly, if the memory has a high bandwidth from memory to the calculation units or very efficient memory management, the speed of the computations could be a bottleneck. 

For FP8 implementations, there is no standard exactly defining what these choices are. Thus, anytime you see a result on networks for FP8 training, it could mean many things. For example, The Nvidia/Arm/Intel whitepaper on FP8 (\cite{nvidia_fp8}) only executes convolutional and linear layers in FP8 for their ImageNet results. All the intermediate representations are stored in FP16, and other operations (including the very expensive Softmax) are executed in FP16. This would generate a significant amount of overhead on efficient inference devices. It is even possible that, in practice, no significant speed-ups are achieved if the network is mostly activation memory bottlenecked, as frequently happens on, e.g. networks that operate on very large images. 

Furthermore, it is likely that networks trained in FP8 will only partially be in the FP8 format. There are currently only two libraries available for training in mixed precision on Nvidia GPUs: The current Apex engine (\cite{apex}), which is also included in PyTorch through the AMP library, and the transformer engine (\cite{transformer_engine}). The Apex library was created to perform faster training, switching between FP32 and FP16 automatically. Depending on the underlying distributions, it will choose the more efficient or the more accurate format. The existence of this logic also indicates the necessity for the FP32 format for some layers during training. The same holds for the transformer engine library. Linear operations are conducted in FP8, but some intermediate activations and operations, such as the Softmax and elementwise adds are done in FP16 instead.

Anytime you see a comparison between number formats, it is essential to have the choices that were made very clear since this strongly affects both the accuracy and the on-device performance of the network. The more is quantized, the more noise is added to the network, and the more the accuracy degrades. 

For INT8, we generally have all intermediate activation values also in INT8. This helps significantly when activation memory/bandwidth is the bottleneck in networks, which happens especially for large image sizes. When quantizing networks for efficient on-device inference, we also have all layers in INT8 or, in the worst scenario, W8A16 for only a limited amount of layers. On top of this, due to integer accumulation being precise, we can make do with a much more efficient accumulator implementation. The effect of this is a significant reduction in energy consumption, as will be discussed in Section~\ref{subsec:hardware_considerations}. 

The above section can be summarized as follows. The proposed FP8 implementation is indeed faster compared to FP16 for networks that are weight, memory, or calculation speed-dominated. However, for networks with large activation tensors, the FP16 activations will still be a bottleneck, and speed-ups will be greatly reduced. The actual speed-ups will also depend on how many layers can actually be executed in FP8 instead of FP16 or FP32. The resulting FP8 mixed-precision networks are more efficient than their pure FP16 counterparts, but a network that is in full INT8 is expected to be significantly more efficient yet. In the next section, we show that on top of this, floating point matrix multiplications are in general less efficient if the actual hardware is considered.

\subsection{Floating point hardware considerations} \label{subsec:hardware_considerations}

Another axis for comparison is the hardware necessary to implement an actual multiply-add. Since most hardware accelerators consist of arrays of multiply-add units, we can analyze a single unit's performance. This informs us of what the picture looks like for the entire array.

We present a simple but fair estimation of the rough area/power cost of both floating-point and integer arithmetic; for two types of accumulators, Kulisch and floating-point accumulators. This analysis assumes that area and power are proportional to the count of equivalent 2-input simple gates (i.e., NAND2/NOR2) needed to construct a design. 
Naturally, a raw gate count is just an indication of area and power. After synthesis and physical design in a given technology, the relative results may vary. However, as a first-order approximation, this measure is strongly correlated with actual performance in practice (\cite{area_estimation}).

As a bit of background, Kulisch (fixed-point) accumulators are both exact and more efficient for number formats like INT8, whereas floating-point accumulators are inexact but more efficient for number formats with a larger number of exponent bits, like \Efour, to allow coverage of the larger dynamic range of producs without increasing accumulator size too drastically. 
Due to the inexact nature of floating-point accumulators, order of summation matters and can yield different results.
Furthermore, the larger the summation, the more inaccurate the accumulated value becomes.
It is unclear what floating-point bit-width is sufficient for the accumulator;
Nvidia's Hopper architecture supports FP16 and FP32 floating-point accumulation (\cite{hopper_in_depth}).

\begin{figure}
\centering
\begin{subfigure}{.5\textwidth}
  \centering
  \includegraphics[width=.7\linewidth]{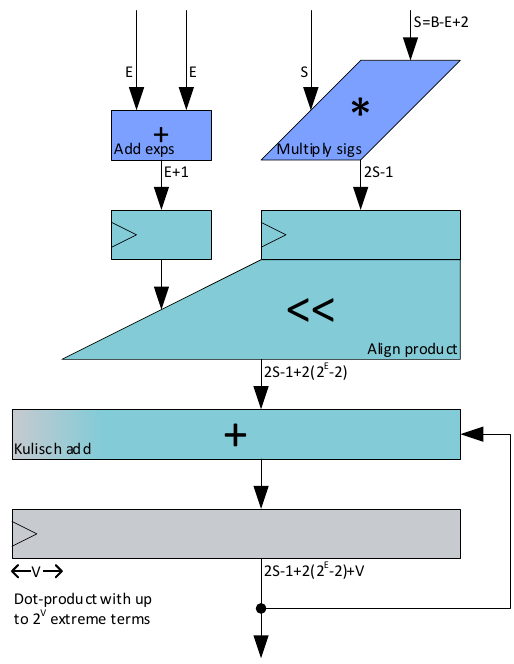}
  \caption{Fixed-point accumulation}
  \label{fig:sub1}
\end{subfigure}%
\begin{subfigure}{.5\textwidth}
  \centering
  \includegraphics[width=.7\linewidth]{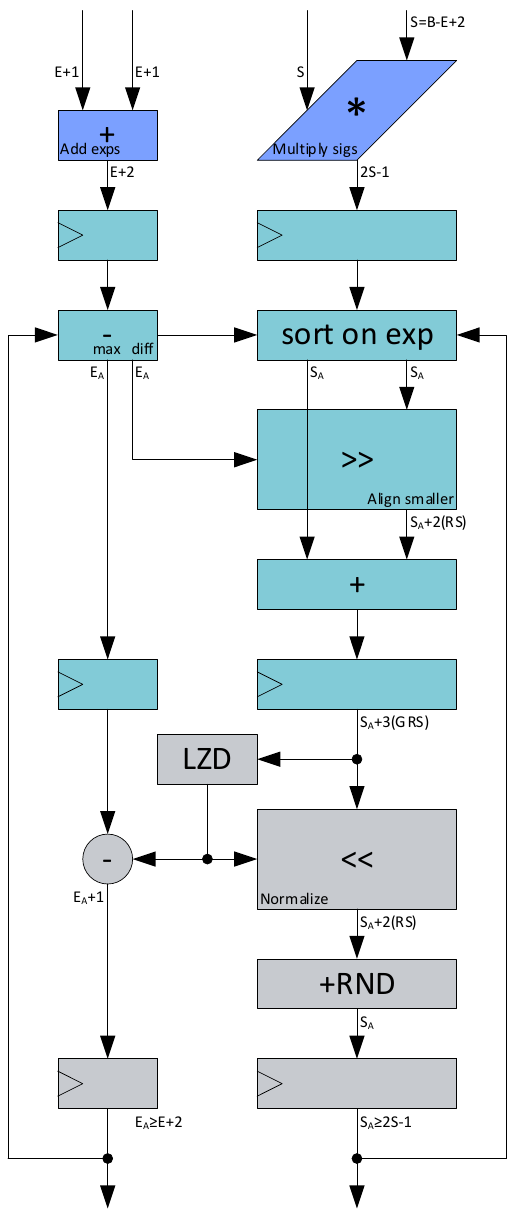}
  \caption{Floating-point accumulation}
  \label{fig:sub2}
\end{subfigure}
\caption{A schematic overview of the components that go in a multiply-accumulate unit in silicon. (Left) is the picture for a fixed-point (Kulisch) accumulator, (Right) for a floating-point accumulator. The dark blue parts represent the logic necessary for the multiplication itself. The grey area is needed for accumulation, and the light blue/green aligns/adds a product to the accumulator.}
\label{fig:hardware_accumulator_comparison}
\end{figure}

In Figure~\ref{fig:hardware_accumulator_comparison}, we present a schematic overview of the hardware for a single multiply-add. On the left, we see the fixed-point and floating-point schemes with a fixed-point accumulator. The analysis is based on \cite{accumulator_estimation}. With fixed-point accumulation and floating-point inputs, a product pipe-stage and product alignment (shifting) are needed. In contrast, integer inputs need neither.

On the right, we sketch the more complicated scheme for floating-point accumulation. Floating-point accumulation adds significant complexity in return for fewer accumulator bits. The design maintains normalization of the accumulator and assumes round-nearest-even. To skew the result in favor of the floating-point accumulator, we do not assume the fewest levels of logic, like a parallel LZA or near/far adder. Instead, we favor the choices for the smallest gate count to make the most optimistic comparison in favor of floating-point.

\begin{figure}[t]
\centering
\begin{tabular}{c}
\includegraphics[width=\textwidth]{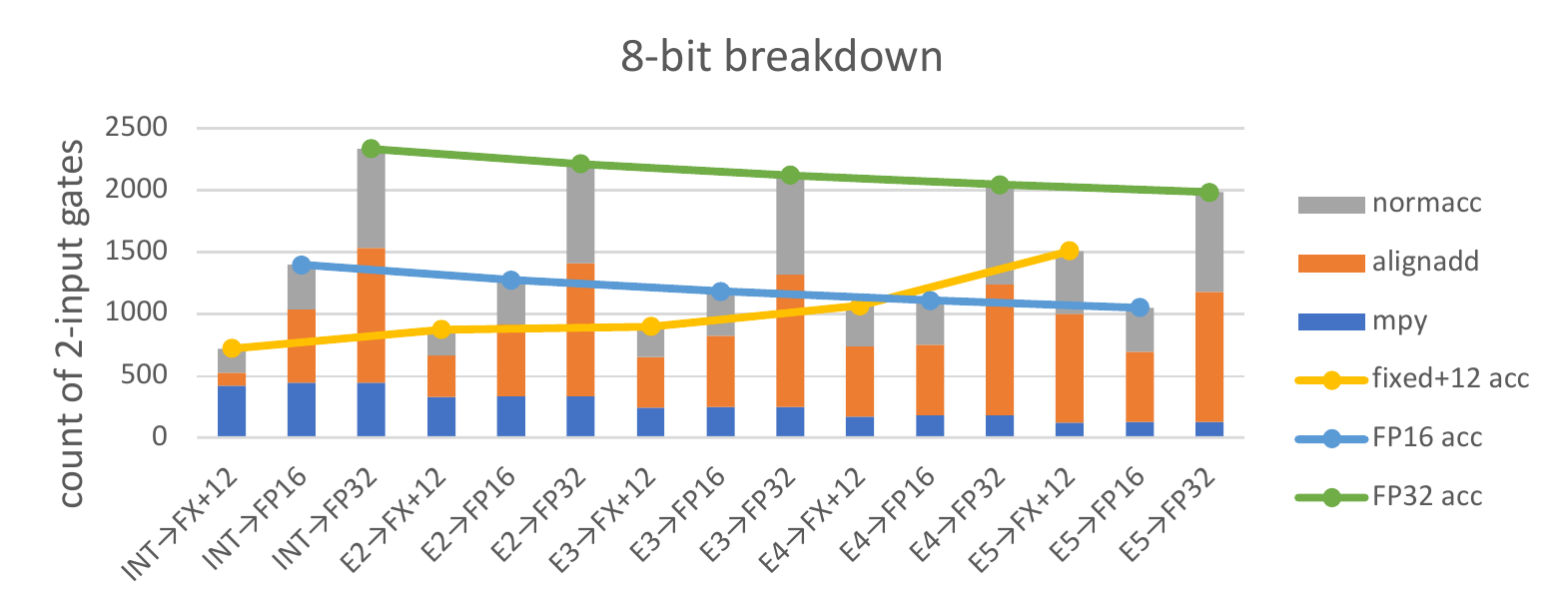}

\end{tabular}
\caption{A count of the number of 2-input gates necessary in hardware to implement each format and accumulator combination. From left to right, INT8 with increasing exponent bits until \Efive. In each group of three, the first bar is for a 15+12=27-bit fixed-point accumulator. The second bar indicates the numbers for FP16 accumulation, and the third bar is the result of an FP32 accumulator. We can see that \Efour->16-bit requires 53\% more gates, and \Efour->32-bit requires 183\% more gates for an implementation in hardware.}
\label{fig:hardware_cost_sketch}
\end{figure}

Based on these schematics, we can count the number of gates necessary to implement both formats. More detailed assumptions are given in Appendix \ref{app:assumptions_gatecount}. See Fig. \ref{fig:hardware_sketch} for the outcome of this analysis. We compare the different number formats and accumulator sizes, with FX+12 indicating the [fixed-point product result]+12 bits to allow for $2^{12}=4096$ extreme products without possible overflow. For INT8, this means a 15+12 = 27-bit accumulator, and for e.g., \Efour, this would be INT37. From this bar-plot, we can deduce the following. For INT8 all the way up to \Ethree, the fixed-precision accumulator is the most efficient. For \Efour the floating-point and fixed-point accumulation almost break even. Most significantly, comparing INT8 with \Efour, there is a 53\% increased cost for the floating-point format. The picture becomes markedly worse if an FP32 accumulator is necessary. Compared to INT8, the \Efour format with an FP32 accumulator is 183\% less efficient. 

Thus, with a dedicated hardware implementation for both formats, \Efour is significantly more expensive than INT8 in area and power necessary for calculations. If networks are bottlenecked by computation, this will also translate to much slower comparative performance. 

 As mentioned, raw gate count is a reasonable first-order approximation of area and power. \cite{microexponents} have done the synthesis and estimate a 40\% decrease in performance of FP8 versus INT8, lending more credibility to our results. Of note is that their results use a floating-point accumulator for both formats, and a fixed-point accumulator would exacerbate the results even further. 

Thus, for efficient on-device inference, the efficacy of floating-point formats falls apart. We have seen in the previous sub-section that for networks that are activation-bandwidth dominated, the floating-point format is worse than INT8 since FP16 is frequently used for the activations. In this sub-section, we conclude that the compute for FP8 is significantly more expensive than for INT8, meaning that compute bottlenecked networks will also see less performance with the floating-point format.

But how about the accuracy? Perhaps FP8 formats are significantly more accurate for deep-learning networks. We will discuss this in the next section.

\section{Deep Learning Network Accuracy Comparison}

In this section, we will provide an in-depth comparison of the accuracy differences between FP8 and INT8 for deep learning network inference in both post-training quantization (PTQ) and quantization-aware training (QAT) settings. We first provide a theoretical lens that explains the difference between the two formats. We show that the only significant difference between the two formats is in their ability to capture outliers. Afterward, we show that this theory predicts performance for both post-training-quantization and quantization-aware training.

We open-sourced the code for reproducing several of our PTQ and QAT results. It can be found at \url{https://github.com/Qualcomm-AI-research/FP8-quantization}.

\subsection{Representing outliers} \label{subsec:representing_outliers}

We first look at the number format itself. As we recall from Section~\ref{sec:preliminaries}, the only difference between INT8 and FP8-EX is that an X number of exponent bits are used for an increased dynamic range instead of accuracy.

\begin{figure}[t]
\centering
\begin{tabular}{c}
\includegraphics[width=\textwidth]{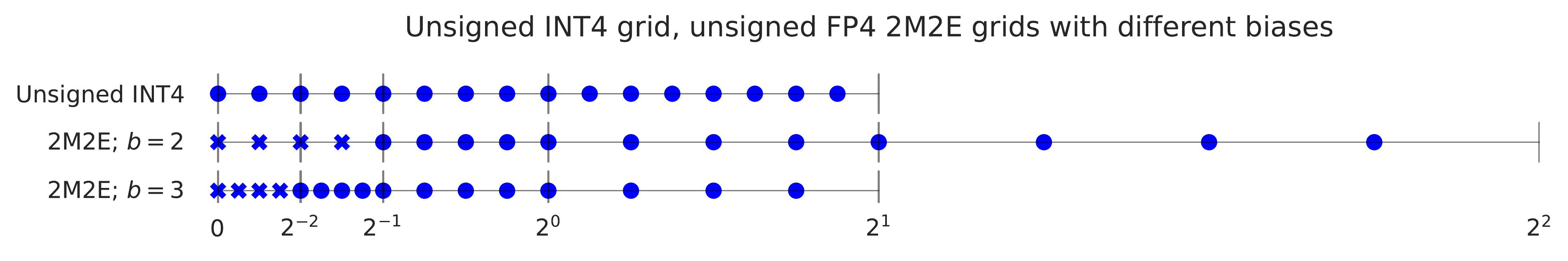}

\end{tabular}
\caption{An example of the floating point format for unsigned 4 bits. When the scaling factor $b$ is flexible, the integer and floating-point format can occupy the same range of representable values. The only difference is in their underlying distributions, where the floating-point format can capture either values closer to $0$ more accurately or represent outliers better. This comes at the cost of the accuracy of the number representation in the other region.}
\label{fig:fp8_number_format}
\end{figure}

It is common in integer hardware to have a flexible scaling factor, either per-tensor or per-channel. This is the same scaling factor as was discussed in Section~\ref{sec:preliminaries}. From a hardware perspective, this is essentially free to implement and improves accuracy significantly, making it commonly used for efficient inference (\cite{krishnamoorthi, whitepaper}). If we take a similar setup for the floating-point format, as is also done by e.g., Nvidia (\cite{hopper_in_depth}), we can see that there is only one difference between the two formats. Adding exponent bits adds more density of values close to zero and less further away from zero. Thus, if the distribution you want to represent has a distribution close to zero and a few outliers, the floating-point format will give less error than the integer format. 
See Figure~\ref{fig:fp8_number_format} for an example.

\begin{figure}[t]
\centering
\begin{tabular}{c}
\includegraphics[width=\textwidth]{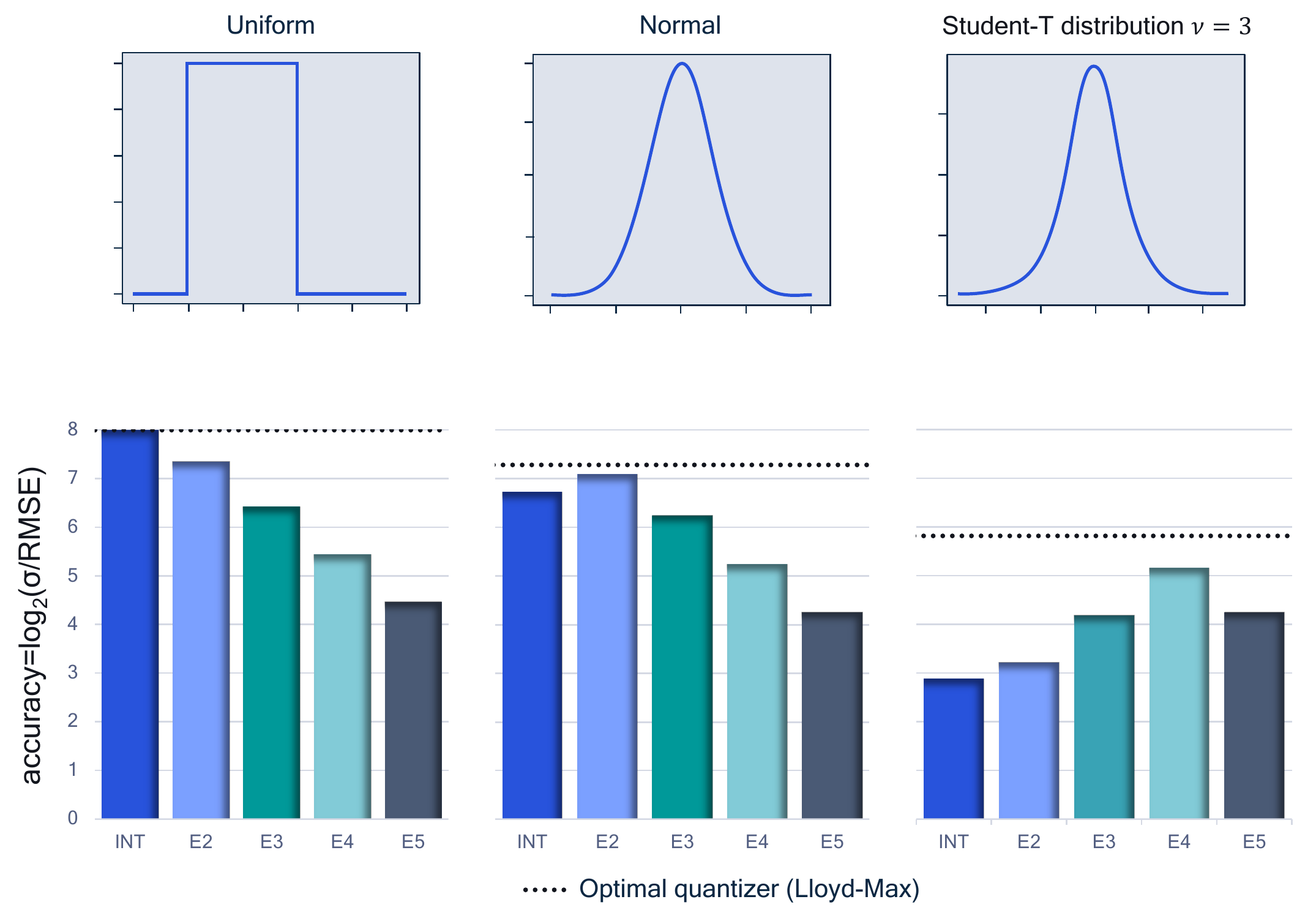}

\end{tabular}
\caption{Here we plot, for several distributions, `bits of accuracy':  inverted and normalized RMSE. More bits of accuracy is better. For the uniform distribution, INT8 is the best. For normal distributions, \Etwo is optimal, with INT8 as a close second. Many distributions in neural networks are Normally distributed, meaning results on this distribution is a very relevant indicator of performance. Only when outliers enter the picture do formats with more exponent bits start giving a better result. The optimal quantizer is the best you could get for these distributions based on the Lloyd-Max quantizer.}
\label{fig:distribution_comparison}
\end{figure}

We can look at a few simple distributions to see what this difference in the distribution of points 
means. We consider the mean-squared error of these distributions, as this has been shown to correlate strongly, both mathematically and practically, with the effect of noise on neural networks (\cite{adaround}). In Figure~\ref{fig:distribution_comparison}, we can see that the INT format is the best for a uniform distribution. This is natural, as the uniformly distributed grid points match the real-valued uniform distribution. For the Gaussian distribution in the middle column, \Etwo is the best, with INT8 as a very close second. This is important to note since many weights and activations in neural networks are well-regularized, either explicitly due to weight regularization or implicitly because of SGD (\cite{rethinking}), meaning that many distributions that occur in practice are quite similar to a Gaussian. 

It is only when outliers come into play that having more exponent bits to match the distribution makes sense. In Figure 5.c., we analyze the MSE of a student-t distribution with significant outliers. We see that the \Efour format performs the best for this specific distribution with outliers.

This leads us to a very simple conclusion: \underline{It’s all about the outliers}. If a distribution has very significant outliers, the \Efour/\Efive format is more accurate, but if the distributions are well-behaved and more Gaussian-shaped, then INT8 or \Etwo/\Ethree is expected to perform better. This is the very simple idea that explains much of the behavior of these formats in practice, as we will see in the following sections. 

It is important to note that adding exponent bits in favor of mantissa bits can hurt accuracy significantly. For well-behaved Gaussian layers, the \Efive format only gives around 4 or 5 bits of information. This means that if the scale factor is set correctly, you could have even used a INT4/INT5 format instead to achieve the same accuracy. 

Lastly, even in the ideal scenario for distributions with outliers, the bit-accuracy is lower than for distributions without outliers. You can see this in Figure~\ref{fig:distribution_comparison} at the bottom. The highest score for the uniform distribution is 8 bits, whereas for the student-t-distribution \Efour has only 5 bits of accuracy. Thus, to optimize your network for quantization, it is best to have a network without any outliers at all. For such a network, the INT8 and \Etwo formats would perform best. This will comes back in Section~\ref{sec:delving_deeper_qat} on QAT. In training networks for quantization, it is easy to train outliers away, resulting in better results overall, and in many cases, the INT8 format then outperforms the floating-point formats. 

\subsection{Setup for comparative analysis}

In the section below, we set out to compare the INT8 formats with the FP8 formats with a different number of exponent bit-widths. We do so by comparing them both in the PTQ and QAT settings. 
In the following PTQ setting, we take an FP32 trained network and convert it without any tricks, ‘naively,’ to the 8-bit formats. The results will not necessarily be indicative of real-world performance, as tricks like AdaRound (\cite{adaround}) can be employed to optimize these networks further. However, we can use these results to compare the formats as fairly as possible.
For the QAT setting, we fine-tuned networks that were trained in FP32. This is more akin to how the FP8 format will likely be used, allowing you to train with it in the loop on a GPU. We will see a remarkable consistency between the theoretical results, the PTQ results, and finally, the QAT results. The QAT setting is not perfectly the same as training networks quantized from scratch, but in our experience the two give nearly identical performance. 

The experimental setup for our comparisons can be found in Appendix \ref{app:experimental_setup}.

\subsection{PTQ Results} \label{sec:ptq}

We consider first the post-training quantization setting. See Table~\ref{tab:ptq} for the results. 

\begin{table}[h!]
    \centering
    \begin{tabular}{l | l|r | r r r r r }
        Task & Model & FP32 $\uparrow$& INT8 & \Etwo & \Ethree & \Efour & \Efive \\
        \hline
         \multirow{8}{*}{Classification}& ResNet18 & 69.72\% & -0.08\% & {\bf-0.06\%} & -0.27\% & -1.15\% & {\underline{-4.8\%}} \\
         & ResNet50 & 76.06\% & -0.07\% & {\bf-0.05\%} & -0.8\% & -0.99\% & {\underline{-3.82\%}} \\
         & EfficientNet B0 & 77.35\% &	{\underline{-14.65\%}} &	-4.78\% &	{\bf-2.8\%} &	-3.7\%	& -9.18\% \\
         & EfficientFormer & 80.21\% & {\underline{-50.35\%}} &	-4.95\% &	{\bf-0.41\%} &	-1.43\% &	-7.41 \% \\
         & DLA102 & 77.94\% & {\bf-0.19\%} & -0.31\% & -0.54\% & -2.13\% & {\underline{-7.95\%}} \\
         & MobileNetV2 & 71.70\% & -0.76\% & {\bf-0.64\%} & -1.08\% & -5.65\% & {\underline{-22.19\%}} \\
         & MobileNetV3 & 73.84\%	& -3.71\% & {\bf-1.48\%} & -1.62\% & -3.33\% & {\underline{-12.06\%}} \\
         & ViT & 77.75\% & -1.33\% & -0.45\% & {\bf-0.04\%} & -0.19\% & {\underline{-76.69\%}} \\ 
         \hline
         \multirow{1}{*}{Object Detection}& YoloV5 & 56.3 & -1.8 &	-0.5	& {\bf-0.3}	& -2	& {\underline{-9.9}} \\
         \hline
         \multirow{6}{*}{Segmentation}& HRNet & 81.05 & -0.12 & -0.02 & {\bf-0.01} & -0.28 & {\underline{-1.06}} \\
         
         & CP-Pointpillar & 40.94 &	{\underline{-21.41}} &	-14.81 &	{\bf-2.86} &	-2.93 &	-7.06  \\
         & RangeNet++ & 0.305	& -1.67	& -0.9	& {\bf-0.4} &	-1.3 &	{\underline{-3.8}} \\
         & FFNet & 79.16 &	{\underline{-1.15}} &	{\bf-0.26} &	-0.31 &	-0.41 &	-1.07 \\
         & DeeplabV3 & 72.91 & -1.67 & {\bf-0.33} & -1.63 & -34.98 & {\underline{-66.38}} \\
         & SalsaNext  & 55.80 & -1.58 & -0.28 & {\bf-0.13} & -0.68 & {\underline{-2.72}} \\
         \hline
         \multirow{1}{*}{NLP}& BERT (GLUE) & 83.06 & {\underline{-12.03}} & -2.75 & -0.45 & -0.26 & {\bf-0.25} \\
         \hline
         \multirow{1}{*}{SuperResolution}& QuickSRNet & 32.79 &	-0.8 & {\bf-0.44} &	-1.78 &	-3.24 &	{\underline{-6.8}} \\

    \end{tabular}
    \caption{Simple PTQ results for many different networks for the INT8 and FP formats with different numbers of exponent bits. Best results for a model are marked bold, worst results are marked with underscore. We see that no format is best for all. Note that all results can potentially be improved with different tricks; results are chiefly for theoretical number format comparisons. }
    \label{tab:ptq}
\end{table}

Looking at these results, there is a clear pattern. For networks like ResNet18 (\cite{resnet}), MobileNetV2 (\cite{mobilenetv2}), and DeeplabV3 (\cite{deeplabv3}), we know that the layers are relatively well-behaved. The layers are mostly Gaussian-shaped. As predicted by our previous theory, the \Etwo and INT formats perform the best,  whereas a format like \Efour and \Efive performs significantly worse. 
We also see that the ViT (\cite{vitdosovitskiy2020image}) and BERT (\cite{bert}) transformer models perform best with \Efour. The reason for this is well understood from the literature: A few layers in transformer networks have significantly large outliers. This was covered in our own paper (\cite{bondarenko}), as well as several other works (\cite{smoothquant, llmint8}). Specifically, there are a few layers in which the activations before the layer-norm have significant outliers. Because these outliers impact performance significantly, leading to $0$ error when clipped, the \Efour format performs best, and the \Etwo/INT8 formats are significantly worse. 

These problems with transformers are very pathological to this specific architecture. They only happen for a very small number of layers and only for a very small number of output channels. We dive deeper into the root cause and solutions in the transformer Section~\ref{sec:transformer}. 

\subsubsection{What do these PTQ results tell us about INT8 versus the \Efour and \Efive formats?}

Similar to our theoretical analysis, the story revolves around outliers. For well-behaved networks without many outliers, the INT8 format is significantly more accurate in the PTQ setting than \Efour and \Efive. We also see that \Efive is never the best format for inference; even for the transformer layers with significant outliers, the \Efour format is better. This is consistent with findings of the Nvidia/Intel/ARM paper (\cite{nvidia_fp8}) that suggest using \Efour for the forward pass and \Efive only for the backward pass. 
Solely looking at the PTQ setting, there is no one format that is best. For some computer vision networks, INT8 is better; for some networks with significant outliers, the \Efour/\Efive formats are better. Purely taking these results into account, the \Efour format looks comparatively worse than \Etwo and \Ethree. Combining these findings with the hardware implementation costs described in Section~\ref{sec:hardware}, the \Efour format itself looks like it is a worse choice than its lower-exponent bit brethren, which are both cheaper hardware-wise and more accurate. If anything, the \Ethree format stands out positively in this accuracy analysis compared to other FP formats. 

\subsection{Quantization-Aware Training}

Arguably the most important scenario for the FP8 format would be what happens when training with the quantized operations in the loop, also referred to as quantization-aware-training (QAT). In this procedure, we take the pre-trained FP32 models and fine-tune them for FP8 quantization. This is very similar to training networks natively in FP8, which would happen on an FP8-equipped GPU. 

For this procedure as well, the experimental details are in Appendix \ref{app:qat}. In summary, both weights and activations are quantized to the FP8 formats. Similar to the PTQ setting, the entire network is quantized to FP8 while training. We perform range-learning based on the LSQ method (\cite{lsq, lsq+}), so that the results are not affected by a difference in setting the quantization ranges. The training setups for INT8 and the different FP8 formats are identical. Both these points allow for the fairest possible comparison between the formats.

\begin{table}[h!]
    \centering
    \begin{tabular}{l|r | r r r r r r}
        Model & FP32 & INT8 & \Etwo & \Ethree & \Efour & W4A8\\
        \hline
         ResNet18 & 69.72 &{\bf 70.43} & 70.25 & 70.20 & \underline{69.35} & 70.01 \\
         MobileNetV2 & 71.70 & {\bf 71.82} & 71.76 & 71.56 & \underline{70.89} & 71.17\\
         \hline
         HRNet & 81.05 & {\bf 81.27} & 81.20 & 81.14 & \underline{81.06}  & - \\
         DeeplabV3 & 72.91 & {\bf 73.99} & 73.67 & 73.74 & 73.22 & \underline{73.01} \\
         SalsaNext (SemanticKITTI) & 55.80 & \underline{55.0} & 55.3 & {\bf 55.7} & 55.2 & - \\
         \hline
         BERT (GLUE avg) & 83.06 & 83.26 & 81.20 & 83.74 & {\bf 83.91} & \underline{82.64} \\

    \end{tabular}
    \caption{ QAT results for the different tested formats. Most results recover performance compared to the baseline FP32 numbers. The best performance per model is marked in bold, worst performance is underlined. Relatively INT8 improves the most compared to the other formats. \Efour is now never best. Results are better for formats with fewer exponent bits as outliers are easy to train away, improving results for those formats more than a format like \Efour. We omitted \Efive experiments as the number format is not considered further for efficient inference.}
    \label{tab:qat}
\end{table}

Let us take a look at Table~\ref{tab:qat}. The first thing to note in the QAT results is that all networks get close to their original floating-point performance. In most cases, we even see an improvement over the baseline FP32 results. The reason for this is simply that training these models for longer generally improves results, even in we would train longer in FP32.

Secondly, we see a similar pattern in the QAT results as we saw in the PTQ setting. For the image-related networks, the INT8 performance is better than the floating-point performance. On top of this, for both DeeplabV3 (\cite{deeplabv3}) and HRNet (\cite{HRNet}), INT8 is also the best format. For the networks like the transformers and SalsaNext (\cite{cortinhal2020salsanext}), where we saw good \Ethree and \Efour performance in PTQ, the results are better in those formats as well. Finally, \Efour is never the best format after QAT. INT8 is a better format for the 2D computer vision networks, and \Ethree is generally better than \Efour for all networks.  

One surprising trend is that the INT8 results improve more than their PTQ baseline than their FP8 counterparts. There is a good reason for this; again, it’s about the outliers. When performing QAT, outliers are clipped and do not receive a gradient. The clipped weights/activations then tend towards the clipping threshold due to regularization. But most importantly, with the outliers clipped, the network learns weights that still perform well despite the outliers being removed. At the same time, when training the quantization parameters with a method like LSQ (\cite{lsq}), the network can learn to make the ranges smaller so as to find a better trade-off between the clipping and quantization errors. The smaller the range, the more accurate your quantized representation will be. This is especially the case for INT8, where the sensitivity to the quantization ranges is much larger than the floating-point formats with more exponent bits that are naturally more resistant to outliers. This way, the INT8 format benefits significantly more from QAT than the FP formats.

The last piece of this puzzle was discussed at the end of Section~\ref{subsec:representing_outliers}. The best achievable MSE for a neural network layer would be when the weights are uniform, and INT8 was used. If the distribution is more Gaussian-like, then \Etwo would be optimal, with INT8 as a close second. The MSE achieved for these well-behaved distributions is significantly higher than what you would get out of the \Efour distribution, even in the most optimal scenario for that format. Thus, if the network could find a weight distribution that is more uniform or Gaussian during training, essentially getting rid of the outliers, the MSE for that layer would be reduced, and we would expect to see a performance improvement. Thus, when training, it is not surprising the INT8 results turn out better than the results for \Efour. The notable exception is the transformer network BERT. We will see in Section~\ref{sec:transformer} that there is a very specific reason for this, as the outliers have some merit for the network itself. 

Finally, for ResNet18, MobilenetV2 and DeepLabV3, W4A8 (4-bit weights and 8-bit activations) is performing better or on par with \Efour. That format would be roughly 50\% more efficient in terms of compute and weight-memory movement. All-in-all, this does not paint a great picture for the floating-point format. 

\subsection{Delving deeper into the QAT networks} \label{sec:delving_deeper_qat}

One thing you might wonder is if the network’s parameters and activation don’t naturally shape themselves around the number distribution that is used. Surely, the gradients will take care of the task of dealing with the number formats, right? 

It turns out that that is not the case. Although it is easy to train away the effect of outliers and learn the quantization ranges, during training, the weight and activation distributions inside the representable grid do not automatically shape themselves around what is optimal. We can look at some results from what happens when we convert two \Efour QAT-trained models back to INT8. In this case, naively, with just simple range setting. As we see in Table~\ref{tab:conversion1}, the accuracy for a network like ResNet18 or MobileNetV2 stays roughly the same or even slightly improves when converting the model from \Efour to INT8. This is despite the network being trained entirely fine-tuned in the \Efour format. For MobileNetV2 the performance stays the same, despite the conversion adding more noise to the network. 

\begin{table}[t]
    \centering
    \begin{tabular}{l|r | r r r r r }
        Model & FP32 & INT8 PTQ & FP8 PTQ & FP8 QAT & -> INT8 & -> W8A16 \\
        \hline
         ResNet18 & 69.7 & 69.55& 68.49 & 69.58 & 69.84 & 69.68\\
         MobileNetV2 & 71.7 & 70.94& 64.22 & 70.81 & 70.82 & 71.07 \\
    \end{tabular}
    \caption{What happens when we take an FP32 network, quantize it with quantization-aware training to \Efour and then naively convert it to INT8? For these two networks, the results stay mostly the same or even improve slightly. This is due to the underlying distribution being better quantized with INT8.}
    \label{tab:conversion1}
\end{table}

This might seem surprising, but the underlying cause is that the distributions of the network, specifically for the activations, do not depend on the number format it was trained in. Rather they depend on the other training settings such as regularization, the optimizer, initialization, etc. The activation distributions are continuous, unlike the weights that are forced to a specific value. These distributions are simply better fit with the INT8 format, causing a slight increase in performance, despite the weights probably incurring more noise. Testing this hypothesis, the last column of Table~\ref{tab:conversion1} has the activations in INT16 instead of INT8, and we see an even further increase in accuracy for ResNet18. This improvement would not happen if the activations formed effectively around the FP8 grid. 

\begin{figure}[t]
\centering
\begin{tabular}{c}
\includegraphics[width=\textwidth]{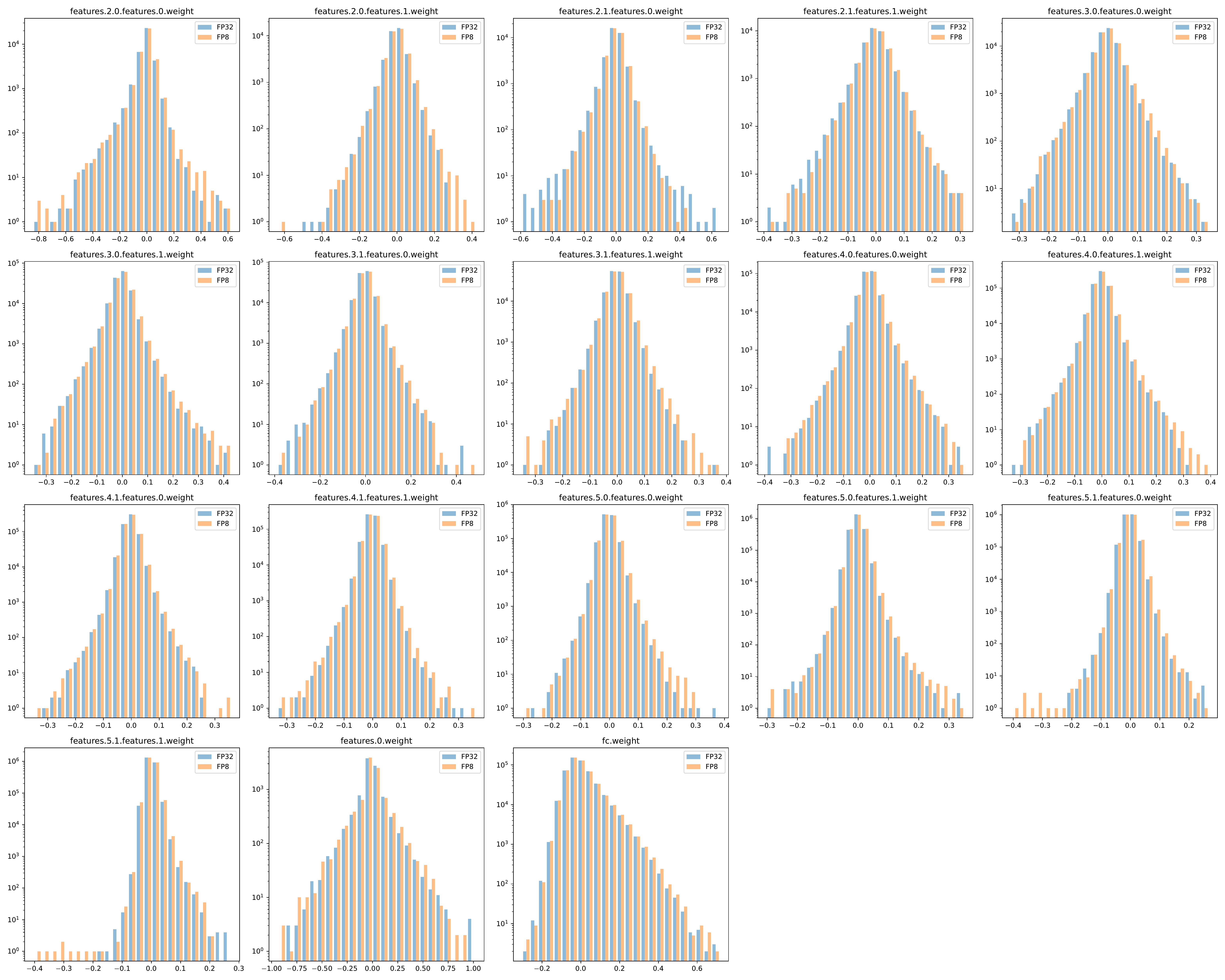}

\end{tabular}
\caption{Distributions of the weights of ResNet18, training from scratch in FP32 (blue) and \Efour (orange). We can see that the different choice in number distribution hardly changes the shape of the distribution. Indeed, not the number format but the other training parameters, such as initialization and the optimizer, determine what weight distributions end up with. }
\label{fig:qat_weight_distributions}
\end{figure}

We can also visualize this for weights. In Figure~\ref{fig:qat_weight_distributions}, we plotted the weight distributions from ResNet18, comparing the weights after training from scratch (from random initialization) for both FP32 and \Efour. We see that the distributions barely differ. Some layers have more outliers for FP32 and others for \Efour. It seems that even the network’s weights don’t really shape themselves around the representational restrictions of the number format.

The above analysis indicates that we can properly talk about number formats being better for representing distributions, similar to the analysis we did in Section~\ref{subsec:representing_outliers}. Weights and activations seem not to learn to match the underlying data format we pick ourselves, but we have to pick the best number format that represents the weights and activations.

\subsection{Transformers} \label{sec:transformer}

So far in our discussion, we have shown that the \Efour format is not better than the alternatives of INT8 and FP8 formats with fewer exponent bits, except for transformers. The problem of transformer quantization is well-known in the literature (\cite{bondarenko, smoothquant, llmint8}), and luckily it is easily fixed in a myriad of ways.

\begin{figure}[t]
\centering
\begin{tabular}{c}
\includegraphics[width=5.0cm]{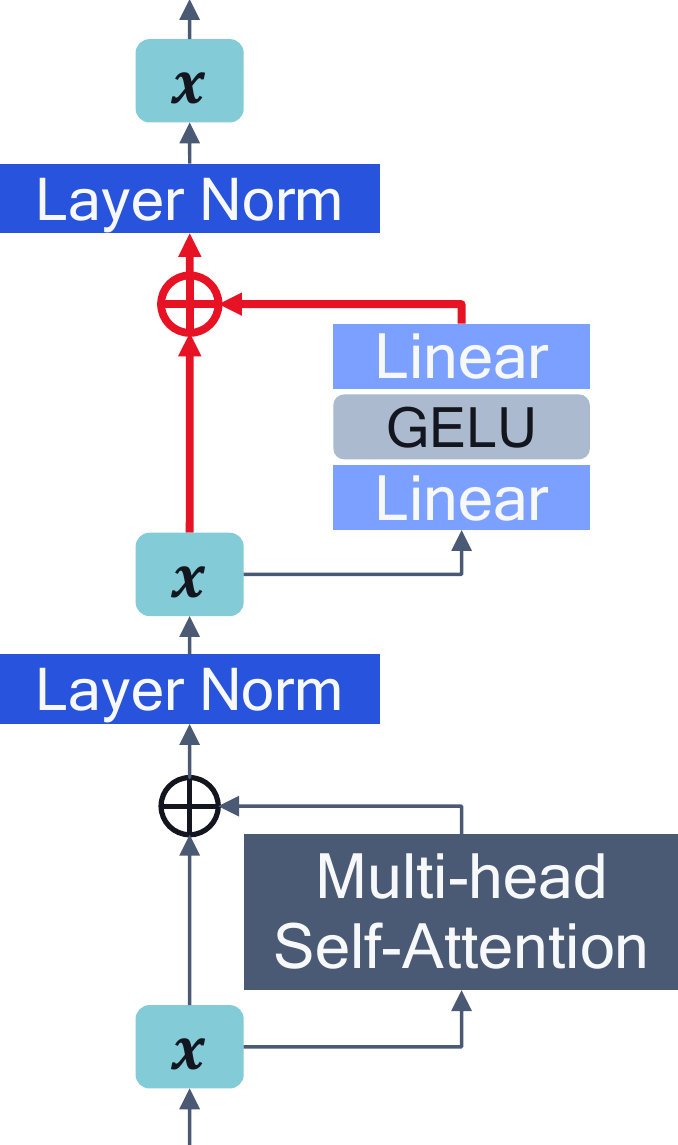}

\end{tabular}
\caption{A single transformer block, consisting of the self-attention module and the linear module. The connection in red is the problematic connection in transformers with outliers. }
\label{fig:transformer_problem}
\end{figure}

The problems with transformer quantization occur in a very specific part of the network, highlighted in red in Figure~\ref{fig:transformer_problem}. There are significant outliers in the summation going into the layer-norm in some of the fully connected modules, especially in the final layers of the network. Simply clipping these outliers reduces the accuracy of the network significantly, as they serve a specific purpose. As described in our paper \cite{bondarenko}, these outliers force the attention mechanism in the next layer to pay attention to some meaningless tokens -- like sentence separator tokens, periods, or commas -- that occur in the text, causing that specific token to not update significantly. In vision transformers, something similar happens, but for meaningless background patches instead. These outliers become more and more pronounced when trained longer.

\begin{figure}[t]
\centering
\begin{tabular}{c}
\includegraphics[width=\textwidth]{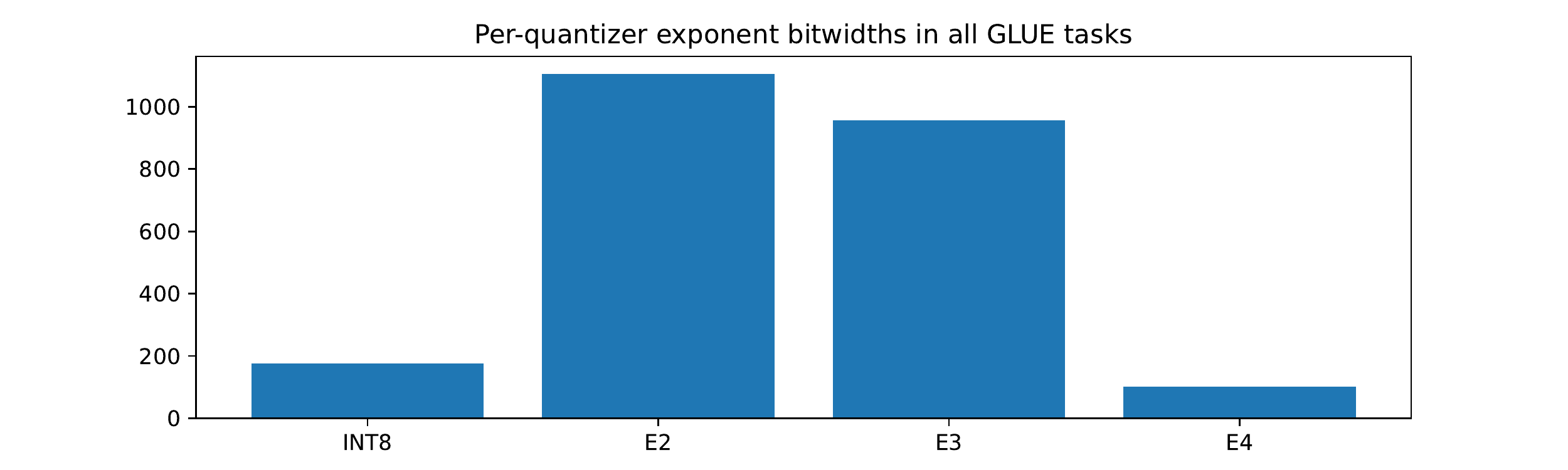}

\end{tabular}
\caption{Optimal PTQ bit-widths for layers in the BERT model for each of the 9 different GLUE tasks. Results are generated by picking the format with the lowest MSE for each layer for that task. Only a few layers are best represented in the \Efour format; similar to the results from our PTQ analysis, most layers perform better in \Etwo or \Ethree.}
\label{fig:transformer_bitwidth_distributions}
\end{figure}

We have plotted the optimal PTQ bit-widths for each quantizer in a BERT model in Figure~\ref{fig:transformer_bitwidth_distributions}. We see that in this analysis, only a few layers are the best in \Efour. The other layers find a lower MSE error with the \Etwo and \Ethree formats. This is a very similar pattern to the PTQ analysis we did in Section~\ref{sec:ptq} for many networks. If the outliers were taken care of in some way for these specific layers that are best in \Efour, we would expect the issues with transformer quantization to disappear.  

Luckily, these outliers are very particular. They occur only in some attention blocks, and in those blocks only in one layer, and in those layers only in a few output channels. These are even the same channels for each data point (\cite{bondarenko, smoothquant, llmint8}).

Because of this, many solutions exist to get excellent accuracy of these networks in INT8 as well, even in the post-training quantization setting that requires little effort. For PTQ, \cite{llmint8} propose having these few dedicated outlier channels run in FP32 instead of INT8. \cite{smoothquant} propose doing a cross-layer equalization procedure like \cite{dfq} in the transformer network, and achieve FP32 results in the INT8 networks as well. \cite{bondarenko}, suggests two possible PTQ changes. (1) a per-embedding group quantization procedure, akin to the per-token quantization proposed in yet another work \cite{zeroquant}; (2), the specific layers with outliers can be executed in W8A16 (8 bit weights, and 16 bit activations) instead of INT8, without a significant drop in accuracy. This last solution is also efficiently supported in existing hardware today. The overhead for this is minimal, as it is only necessary for a small number of layers. 
Finally, a recent paper by \cite{gptq} gets excellent performance with 4-bit weights, also in a PTQ setting, by employing an algorithm that operates in the same vein as AdaRound (\cite{adaround}).

By now, it should come as no surprise that quantization-aware training also recovers the accuracy as detailed in many papers such as \cite{q8bert}. Our paper (\cite{bondarenko}) even recovers much of the original accuracy with 4-bit weights and 8-bit activations. This last result elucidates further why \Efour works well in the PTQ setting. The networks are relatively overparametrized and can take a hit in the representational accuracy of the weights, as long as the outliers are captured sufficiently. Recently, \cite{scalinglaws} have even shown that 4-bit weights are also likely to be the optimal point in terms of size and performance trade-offs for large language models. 

The final verdict on transformers is that, yes, in the most naive PTQ setting, the \Efour format performs better than INT8. However, this happens due to a very specific peculiarity. An issue that can be dealt with in numerous ways. Several works have shown that even in the INT8 PTQ setting you can get back your original accuracy with any of several possible tricks, and quantization-aware training can recover all accuracy as well. Since choosing to execute transformers in \Efour leads to a suboptimal solution in hardware, as discussed in Section~\ref{sec:hardware}, it is likely worthwhile to adopt one of the tricks to fix the outlier issues instead of resorting to a different number format. This way, even 4-bit weights are possible and seem to be optimal.

\section{Comparison to other work}

We compare our results to two mainstream papers that discuss the \Efour/\Efive formats that are being considered, one paper from Graphcore (\cite{graphcore}), and a paper from Nvidia, Arm, and Intel (\cite{nvidia_fp8}). Note that these papers focus more on the training side, whereas we exclusively focus on the impact of efficient inference. We find that the results of both these papers agree entirely with the results in this whitepaper, and the three are entirely consistent. However, in this whitepaper, we presented a more thorough comparison between the formats with a proper theoretical understanding. The parts that paint a different picture for the FP8 format in our whitepaper are missing from both other papers in this area.

The paper from Graphcore only compares INT8 with \Efour/\Efive for a ResNet-32 model on CIFAR-100. Their conclusion for the weight and activation quantization is that both formats can achieve the same performance on the test case, but the set of quantization parameters that the FP8 formats get a high accuracy for is larger. That is generally not a problem for inference, as the tools for setting the ranges are not complex and do a great job of finding correct parameters. The paper chooses to have an integer scaling factor, as opposed to the more common higher-bit floating-point-like scaling factor; the latter would likely skew performance favorably to the INT8 performance. Similarly, no per-channel quantization is considered, which would be the default for integer quantization. The problems with the quantization of gradients put forth in the paper are well-known and have been addressed in many works in the past (\cite{sun2019_HFP8, Gupta2015}). Many works have shown the necessity of stochastic rounding and proper range setting for the backward pass that alleviate these issues and make INT8 for gradients work just as well (\cite{sun2019_HFP8, hindsight, wageubn}).
However, INT training is outside the scope of this paper.

The rest of this paper shows that one can get close to the original FP32 accuracy by training with the format. This is not entirely surprising, as even W4A8, 4-bit weights, and 8-bit activations gets very close to the FP32 accuracy for the considered ImageNet and Language models, and INT8 recovers performance for the tested networks with QAT as well (\cite{whitepaper, lsq, lsq+}).

The Nvidia/Arm/Intel paper (\cite{nvidia_fp8}) is a bit more exhaustive but still omits several of the comparisons that can be found in this whitepaper. This paper also shows that for common image-classification networks, training with \Efour can get back close to the original FP32 training accuracy. This is not surprising for the same reason as stated before, and these results only consider the linear and convolutional layers in FP8, whereas the rest of the network is in FP16. 

The paper also has results for training large language models where networks get back to the original accuracy. We have seen in our transformer Section~\ref{sec:transformer} that these models can be executed in INT8 as well, both with PTQ and QAT. Finally, the only comparison with the INT8 format comes in the form of comparing transformer-based language models in the PTQ setting. This is consistent with our results in Section~\ref{sec:ptq}. However, as argued, these problems are easily fixable for transformer networks, making them able to execute entirely in INT8 or in mixed precision with a small number of activations in INT16. Results similar to the PTQ comparisons where we show INT8 outperforms \Efour are not found in this paper. 

There is very limited discussion in the paper on the experimental setup and the quantization choices that were made, making a thorough comparison and discussion very difficult and an in-depth analysis impossible. Nevertheless, we can make some estimated guesses. In the section on computer vision models, the paper indicates they keep the activations in FP16. As discussed in Section~\ref{sec:hardware}, this would not be preferable for efficient on-device inference, especially on large-image sizes, and could significantly reduce performance. Keeping intermediate activations in FP16 would, however, significantly improve the accuracy of quantized networks, as the quantization of operations like element-wise-additions, pooling-operations, and non-linearities does not introduce extra noise into the network. A more apt comparison would be if the format was compared to INT8 with INT16 activations in terms of accuracy. 

There are other works that discuss the FP8 format, such as \cite{wang2018_fp8_training, sun2019_HFP8}, but they do not have a comparative analysis to other formats, and the results are limited. Similarly limited are the results in \cite{Huang2021_8bit_flex_fp}, but the paper shows for a single network that the \Etwo format for weights and \Ethree format for activations is best on the VGG network. All of these previous works leave the hardware considerations out of the picture. One paper that does have a very extensive hardware evaluation of the formats on an extensive suite of models is \cite{microexponents}. The paper actually does a synthesis of the hardware for INT and FP and shows that our analysis from Section~\ref{sec:hardware} holds, indicating a 40\% performance decrease of FP8 compared to INT8. 

In conclusion, our results agree with, and expand on, the results from the cited papers. By showing a broader set of results, we show a lot less rosy picture for the \Efour/\Efive formats. In direct comparison, there are no significant performance gains from FP8 over the standard INT8 format.

\section{FP8 to INT8 network conversion}

First, let us see what happens when \Efour trained networks are converted naively to INT8. The full experimental setup for this is given in Appendix \ref{app:experimental_setup_conversion}, and a brief overview of this experiment is presented here.

To simulate training on FP8 hardware, we perform FP8 QAT on various models. We train ResNet18 and MobileNetV2 using FP8 QAT from scratch to ensure that potential differences in weight or activation distributions caused by FP8 quantization are captured in our experiments. However, as discussed in Section~\ref{sec:delving_deeper_qat} we find no significant differences in the learned weight distribution, so for the other models, we use the FP8 QAT fine-tuned FP32 models presented in Section~\ref{sec:delving_deeper_qat}. After QAT, we quantize the learned FP8 weights to INT8 and run inference on the task dataset. We evaluate these models using W8A8 and W8A16 quantization.

\begin{figure}[t]
\centering
\begin{tabular}{c}
\includegraphics[width=\textwidth]{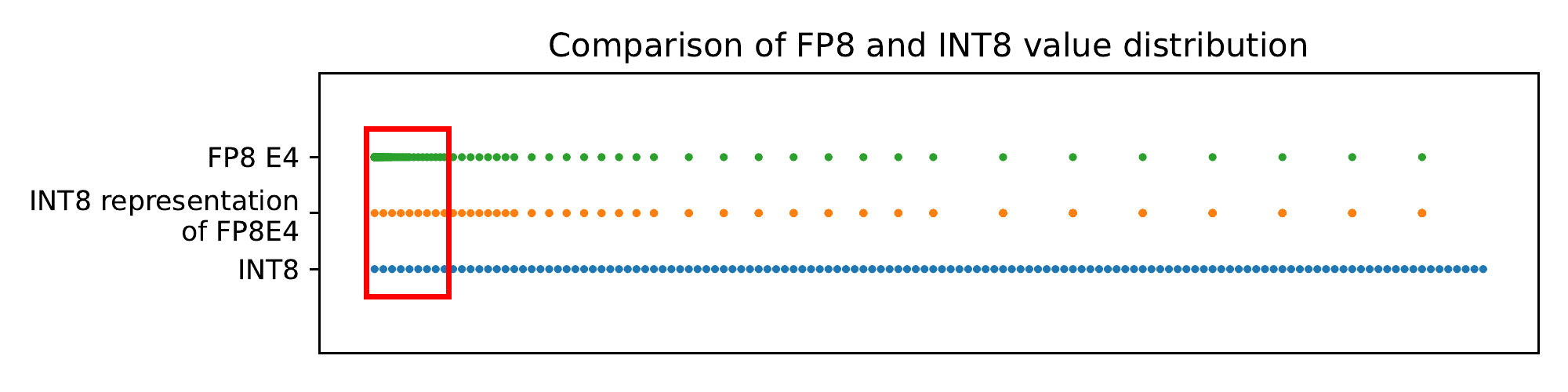}

\end{tabular}
\caption{The number distributions upon conversion. On the top, we see the \Efour number format spread. On the bottom, the INT8 number representation. The middle values are what happens when \Efour is cast. When the scale factor can be arbitrarily chosen, the INT8 number format can exactly match the numbers in the \Efour format for the large values. Only the values in the red box, the smaller numbers, will suffer some loss.}
\label{fig:number_format_overlap}
\end{figure}

As shown in Figure~\ref{fig:number_format_overlap}, INT8 can exactly represent roughly 90\% of the range covered by the \Efour format without any quantization error. As long as the scaling factor can be changed arbitrarily, this match is exact. The remaining 10\% of the range incurs small quantization errors. This has implications for range estimation. As the maximum value of the best matching INT8 grid is larger than the maximum of the target FP8 distribution, using min-max range estimation on FP8 quantized tensors will yield a slightly misaligned INT8 quantization grid. Therefore, we use MSE range estimation for both weights and activations.

\begin{table}[]
    \centering
    \begin{tabular}{l|r | r r r r r }
        Model & FP32 & INT8 PTQ & FP8 PTQ & FP8 QAT & -> INT8 & -> W8A16 \\
        \hline
         ResNet18 & 69.7 & 69.55& 68.49 & 69.79 & 69.84 & 70.00\\
         MobileNetV2 & 71.7 & 70.94& 64.22 & 71.59 & 71.59 & 71.54 \\
         BERT (GLUE avg) &  83.06 & 71.03 &83.08 & 83.91 & 79.38 & 83.54 \\
         SalsaNext (SemanticKITTI) & 55.8 & 52.8 & 55.1 & 55.2 & 54.1 & 54.2 \\
    \end{tabular}
    \caption{What happens when we take an FP32 network, quantize it with QAT to \Efour and then naively convert it to INT8? For a network like ResNet18, the results improve, and for MobileNetV2 the performance stays the same. Only for networks where we saw better performance in PTQ, due to \Efour capturing the outliers, does the conversion degrade performance. }
    \label{tab:conversion2}
\end{table}

Results for these experiments can be found in Table~\ref{tab:conversion2}. For ResNet18 and MobileNetV2, we somewhat surprisingly find that integer quantization slightly improves results compared to the \Efour QAT results. This effect was also described in Section~\ref{sec:delving_deeper_qat}. We attribute this difference to the fact that these models have no significant outliers in activations, and for most of the range representable by FP8 \Efour, INT8 has more precision, as illustrated in Figure~\ref{fig:number_format_overlap}. As a result, INT8 quantization of activations without outliers slightly improves results.

For BERT-base on GLUE and SalsaNext on SemanticKITTI, we see that INT8 quantization decreases accuracy compared to FP8 QAT. However, the degradation is less than when quantizing an FP32 model to INT8. To assess whether accuracy can be improved, we perform INT8-QAT on BERT-base and find that we can recover accuracy up to 83.79. This result is fully expected, as the said networks have outliers. These outliers cause problems when converting from FP32 directly to INT8, and the intermediary of FP8, although improving results for INT8, does not get rid of the outliers entirely, as they can be represented better by the FP8 grid.

The general conclusion is that for networks that were easy to quantize from FP32 to INT8, the conversion is expected to be smooth and can, in several cases, be done directly with simple post-training quantization (PTQ) techniques such as range setting. 
For networks that were already problematic to convert to INT8 from FP32 with simple PTQ techniques because of outlier problems, similar issues will arise when converting to INT8 from FP8. 
However, since these latter networks are trained to deal with the reduced precision of the FP8 format, the INT8 simple conversion results from FP8 are better when compared to INT8 simple conversion from FP32. Moreover, INT8 quantization-aware training (QAT) can be further employed to recover more accuracy in such cases.

\section{The INT quantization paradigm}

As evident from the previous section, the INT8 format can get just as good or better accuracy than the FP8 format while being much more efficient in hardware. The literature on this in the research community is vast and quite exhaustive. Many networks can be quantized with PTQ techniques to INT8 without much of a drop in accuracy. For the remainder of the networks, having W8A16 mixed-precision for some layers solves most quantization issues in PTQ. And the worst-case scenario, with quantization-aware training techniques, it is very rare that a model can not be reproduced accurately in INT8. Appendix \ref{app: aimet_results} has a long list of models that were successfully converted to INT8 with our own AIMET quantization tool (\url{https://github.com/quic/aimet}, \cite{aimet}). 

On top of INT8, the INT4 format has become very popular for the weights in a neural network. This gives an even better trade-off between accuracy and efficiency. Many works (\cite{adaround, brecq, lsq, whitepaper}) have shown the efficacy of doing INT4 quantization for the weights, in both the PTQ and QAT settings. Even for transformers, recent papers show that INT4 is likely the best accuracy/efficiency trade-off for the weights. \cite{scalinglaws} and \cite{gptq} show that large-scale language models can be converted to 4-bit weights even with simple PTQ methods. Appendix \ref{app: aimet_results} also lists models we have tested W4A8 quantization on in AIMET. 

\begin{figure}[H]
\centering
\begin{tabular}{c}
\includegraphics[width=11.0cm]{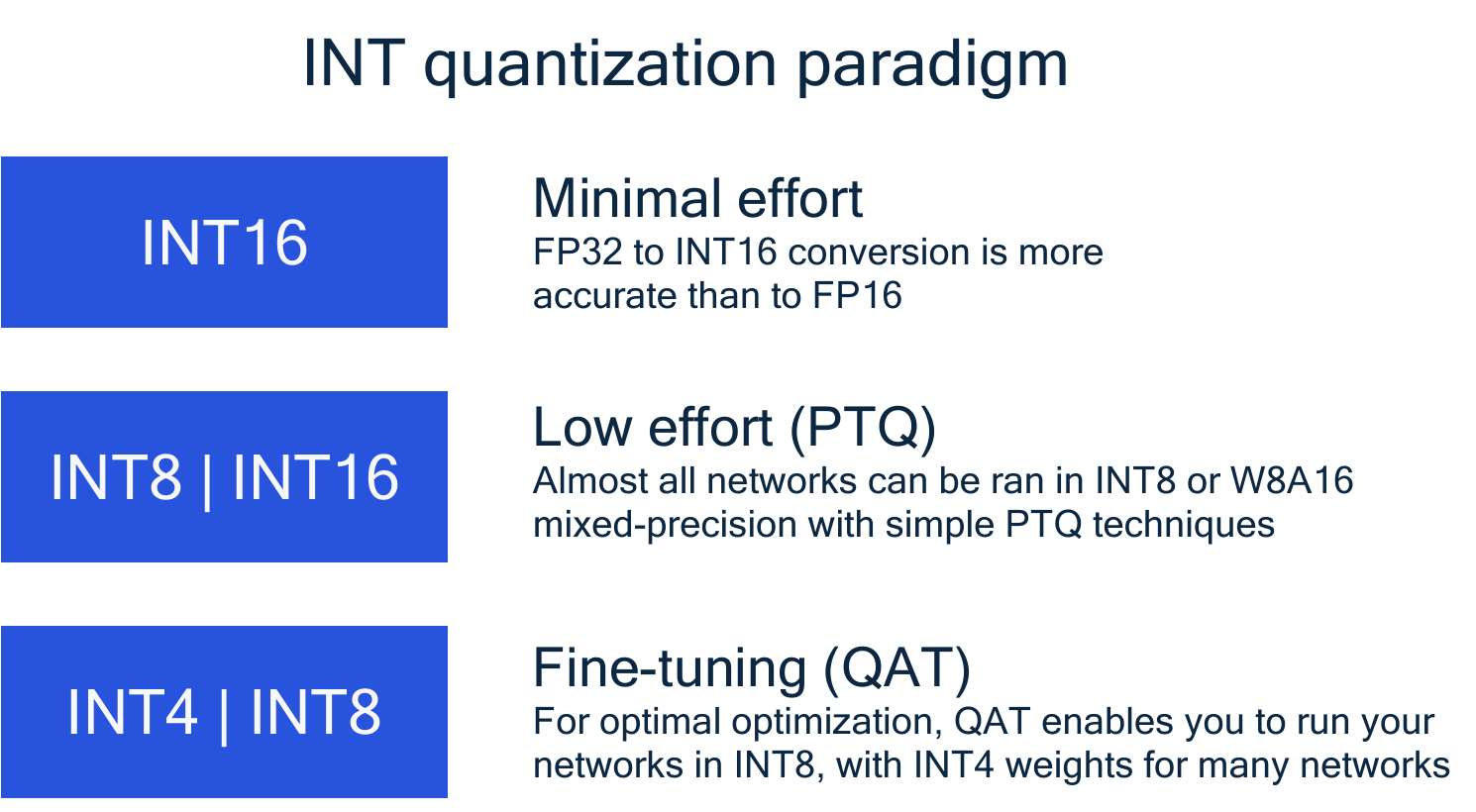}

\end{tabular}
\caption{The INT quantization paradigm.}
\label{fig:int_paradigm}
\end{figure}

This brings us to the INT paradigm in Figure~\ref{fig:int_paradigm}. The INT16 format is the most accurate; it is even more accurate than FP16 for representing FP32 values. If you do not care much about efficiency and just want to deploy without having to take care of quantization, the INT16 format is your best bet. With a low amount of effort, by using PTQ techniques, you can frequently get your networks in full INT8. For some networks, some layers might need more accuracy, in which case W8A16 layers almost always solve the issue. Finally, if you want to really optimize your networks, going with quantization-aware training can get your networks into the 4-bit weight and 8-bit activation regime. This is very achievable for a wide range of networks. This jump in efficiency from 8 to 4-bit weights is very important for weight-bounded networks like the extremely large language models that are being released today. This is an efficiency boost that currently does not exist in the floating-point world. 

The tools for integer quantization have been around for quite some time. Every hardware vendor has some form of integer optimization library for inference. Qualcomm has the AIMET toolkit (\url{https://github.com/quic/aimet}, \cite{aimet}), which is a mature toolkit that optimizes quantization specifically for Qualcomm devices.

\section{Conclusion}

Overarchingly, we have seen that the floating-point formats \Efour and \Efive are not replacements for INT8 for deep learning inference in terms of performance and accuracy. For most well-behaved layers and networks, these formats are worse than INT8, especially when networks are trained with either format in the loop. For corner-case scenarios in PTQ, where layers have significant outliers, the floating-point formats can be better in terms of accuracy. However, there are more efficient solutions for the problematic transformer layers. You can either run these layers in W8A16 with mixed-precision or apply quantization-aware training. 

We have also seen that implementing the FP8 formats in hardware for inference is not efficient and incurs significant overhead. Depending on the accumulator size, the FP8 MAC units are 50\% to 180\% less efficient than their INT8 counterparts. This would make a dedicated chip significantly slower if the workloads are compute-bound. 

Lastly, many networks can easily be quantized into INT8 or even pushing INT4 for even further improved efficiency. The tools for this have been available for many years, and the leap to 4-bit weights is something the floating-point formats do not do as of this writing. 

Because of these reasons, implementing floating point formats for edge use-case scenarios is sub-optimal compared to the standard stack of integer solutions available today. If you want the best accuracy and efficiency trade-off for your models, quantizing them to INT4-INT8-INT16 is the best solution.

\bibliographystyle{icml2020}
\bibliography{references}

\newpage
\appendix

\section{Floating-point number description} \label{app:number_description}

The floating-point number format used in the paper more complicated than the sketch in Section~\ref{sec:preliminaries}. Here we provide a more complete and formal description of the format. 

A floating-point number set $F \subset \mathbb{R}$ is a set whose elements are defined as follows:
\begin{equation}
f=(-1)^s 2^{p-B} (1+\frac{d_1}{2}+\cdots \frac{d_m}{2^m} )
\end{equation}

Where $s \in {0,1}$ is the sign bit, $d_i \in {0,1}$ is the m-bit significand, $p \in N ; p<2^e$ is the $e$-bit exponent, and $B$ is an integer exponent bias, commonly defined to be $2^{(e-1)}$.

Note that this definition does not allow for a representation of $0$.
To allow 0 to be represented, the exponent value $p=0$ is reserved to indicate subnormal numbers. In this case, the exponent value is implicitly set to $1$, and $f=(-1)^s 2^{1-b} (0+\frac{d_1}{2}+\frac{d_2}{2^2} + \cdots \frac{d_m}{2^m} )$

In the paper we make the following further assumptions about the FP8 format:
\begin{enumerate}

  \item  Implicit mantissa bit. The format contains an implicit mantissa bit, which is $0$ when $e=0$, and $1$ otherwise. The mantissa bits that are explicitly stored encode the fractional part of the mantissa; e.g. a 3-bit mantissa value of $111$ encodes a mantissa value of $1.111$ if $e > 0$ and $0.111$ otherwise
  \item  Subnormal numbers. A floating point number with a (stored) exponent value of $0$ indicates a subnormal number. In this case, the number can be decoded with an implicit mantissa bit of $0$ and an exponent value of $1-B$. The subnormal numbers provide uniform quantization around zero, in the range $[-2^{2-B}, 2^{2-B}]$
  \item  Encoding of special values. We assume that binary value $10000000$ is reserved to encode a special value (e.g. NaN) and assume no special encoding for INF. Similar to INT8 quantization we ignore overflow and clip values outside of the FP8 range to the maximum or minimum FP8 value, whichever is closest. This leads to $255$ unique FP8 values, and $1$ reserved value.
\end{enumerate}

Point 1 and 2 imply that formats with $0$ or $1$ exponent bits are (nearly) identical to INT8 format, encoded in sign + magnitude format.

\section{Gate Count assumptions} \label{app:assumptions_gatecount}

These are the assumptions made for the gate-count in our hardware analysis in Section~\ref{subsec:hardware_considerations}

\begin{center}
\begin{tabular}{|l | l |}
\hline
  Component   &  $C=\sum(F_{in}-1) = \sum n_{FI2}$  \\ \hline \hline
   [\ N ]\ AND2 /  [\ N ]\ OR2 & 1 \\ \hline
   AO[\ I ]\ 21 / OA[\ I ]\ 21 & 2 \\ \hline
   MUX2 / X [\ N ]\ OR2 / HA & 3 \\ \hline
   FA & $2C_{HA}$ + $C_{OR2}$ \\ \hline
   FF (non-scan) & $2C_{MUX2}$ \\ \hline
   N-bit add & $(N-1) C_{FA} + C_{HA}$ \\ \hline
   N-bit increment & $N C_{HA}$ \\ \hline
   N*N PP bits & $N^2 C_{AND2}$\\ \hline
   N*N PP reduce & $(N-1)^2 C_{FA} + (N-1) C_{HA}$\\ \hline
   N-bit << [\ 0, k ]\ & $N [\ \log_2 (k+1) ] C_{MUX2} - k(C_{MUX2} - C_{AND2})$ \\ \hline
   N-bit >> [\ 0, k ]\ signed & $(N-1) [\ \log_2 (k+1) ] C_{MUX2}$ \\ \hline
   N-bit << [\ 0, k ]\ widen & $(N-2) [\ \log_2 (k+1) ] C_{MUX2} + k(C_{MUX2} + C_{AND2}$ \\ \hline
\end{tabular}
\end{center}

\section{Experimental Setup PTQ/QAT} \label{app:experimental_setup}

\subsection{Tasks, Datasets, Models, and Metrics}

Below is the list of models used in this study, accompanied by the used dataset description, the metrics used and a link to the original source.
The setups described here are used for both PTQ and QAT experiments, unless otherwise noted.

\subsubsection{ImageNet Classification}

We run classification experiments on ImageNet (\cite{imagenet}). 
For ImageNet experiments we report top-1 accuracy on the validation set, as is usual.

\begin{center}
\begin{tabular}{|l | l |}
\hline
  Model   &  Model Source  \\ \hline \hline
   ResNet18 (\cite{resnet}) & \cite{torchvision}\\ \hline
   ResNet50 (\cite{resnet}) & \cite{torchvision}\\ \hline
   EfficientNet B0 (\cite{effnetB0}) & \cite{effnetB0repo} \\ \hline
   EfficientFormer (\cite{effformer}) & \cite{effformerrepo} \\ \hline
   DLA102 (\cite{dla102}) & \cite{dla102repo} \\ \hline
   MobileNetV2 (\cite{mobilenetv2}) & \cite{tonymodel} \\ \hline
   MobileNetV3 (\cite{mnv3}) & \cite{torchvision} \\ \hline
   ViT (\cite{vitdosovitskiy2020image}) & \cite{timm}\\ \hline

\end{tabular}
\end{center}

\subsubsection{Yolov5}
We run object detection experiments on COCO (\cite{cocodataset}) using Yolov5 (\cite{yolov5}). We start from a pre-trained FP32 model. Model definition and pre-trained parameters from \cite{yolov5repo} repository are used. We report mAP on the COCO dataset.

\subsubsection{GLUE Benchmark}
We run experiments for the 8 standard GLUE (\cite{wang2019glue}) subtasks (COLA, MPRC, SST2, RTE, STSB, QQP, MNLI, and QNLI; excluding WNLI). We use a pre-trained checkpoint for BERT-base (uncased, 109M parameters, \cite{bert}) from the HuggingFace repository (\cite{huggingface}), which we fine-tune on each individual subtask, c.f. \cite{bondarenko}. The GLUE metric is the average of the subtask scores.

\subsubsection{Pascal VOC Semantic Segmentation}
We run semantic segmentation experiments on Pascal VOC (\cite{everingham2015pascal}) using DeepLabV3 \cite{deeplabv3} with a MobileNetV2 backbone. We start from a pre-trained FP32 model. Model definition and pre-trained parameters from \cite{deeplabrepo} are used. We report mIoU on the Pascal VOC test set.

\subsubsection{Cityscapes Semantic Segmentation}
We run semantic segmentation experiments on CityScapes (\cite{Cordts2016Cityscapes}) using HRNet (\cite{HRNet}) and FFNet (\cite{ffnet}). We start all our experiments from a pre-trained FP32 model. Model definition and pre-trained parameters from \cite{hrnetrepo} and \cite{ffnetrepo} are used for HRNet and FFNet, respectively. We report mIoU on the CityScapes test set. 

\subsubsection{Semantic-KITTI point cloud segmentation}
We run point cloud segmentation experiments on the Semantic-KITTI (\cite{semantickitti}) using SalsaNext (\cite{cortinhal2020salsanext}) and RangeNet++ (\cite{rangenet}). We start all our experiments from a pre-trained FP32 model. Model definition and pre-trained parameters from \cite{salsanextrepo} and \cite{rangenetrepo} are used for SalsaNext and Rangenet++, respectively. We report mIoU on the Semantic-KITTI validation set.

\subsubsection{CenterPoint-Pillar}
We run 3D object detection experiments on Nuscenes dataset (\cite{nuscenesref}) using CenterPoint-Pillar (\cite{centerpp}). We start all our experiments from a pre-trained FP32 model. Model definition and pre-trained parameters from \cite{centerpprepo} are used. We report mAP on the Nuscenes dataset. 

\subsection{PTQ setup} \label{app:PTQ}
In our PTQ setup we aim to keep the experimental setups for FP8 and INT8 PTQ as close as possible. Several techniques to improve INT8 PTQ results are ignored in this work, such as CLE, Bias correction or AdaRound (see \cite{whitepaper} for details), as no FP8 implementation for these methods currently exists. Note that as a result, our INT8 PTQ results are lower than those reported in previous works. However, since we do not apply any PTQ techniques for FP8 quantization, we believe that this approach provides the fairest comparison between FP8 and INT8 performance. 

For both INT8 and FP8, we perform a grid search over the following PTQ configurations:
\begin{itemize}
\item Per-channel vs per-tensor quantization
\item Range setting heuristic: MSE and minmax for weights; MSE, allminmax and running minmax for activations
\item With and without batch norm updating as described in \cite{oscillations}, as this technique is agnostic to the quantization method used.
\end{itemize}

For FP8 we additionally experiment with omitting range estimation, and using a fixed range (or bias value), as described in \cite{graphcore}. However, even with an extended search over possible ranges, using a fixed range/bias for a full network always underperforms compared to a per-tensor or per-channel range.

The results in Table~\ref{tab:ptq} are the best results for each network over these hyperparameters.

\subsection{QAT setup} \label{app:qat}
In all QAT experiments we start from a pre-trained model and fine-tune this using the FP8 quantizers from \cite{fp8paper}. In all experiments we keep the number of mantissa/exponent bits fixed. In all experiments, we experiment with both per-channel and per-tensor quantization and report best results. Where applicable we re-estimate batch norm parameters before validation, as described in \cite{oscillations} as this always improved results.

\subsubsection{ImageNet}
In all ImageNet QAT (\cite{imagenet}) experiments we run experiments with and without range learning and with per-tensor or per-channel quantization. For all models and exponent bitwidths, per-channel quantization with range learning performed best. For each combination of model and exponent bitwidth we use the weight and activation range estimation methods that gave best results in PTQ. We search over a wide range of learning rates, between 1e-2 and 1e-6 and report results for the best learning rate. During training we use cosine learning rate decay to 1e-3 of the starting learning rate. In all experiments we use the Adam optimizer for model parameters and the SGD optimizer without momentum for the quantization range parameters. We omit momentum in quantization range optimization as we found that this can cause the range to over- or undershoot. We do not use weight decay. We train all models for 20 epochs using a batch size of 32. 

\subsubsection{BERT/GLUE}
We run QAT experiments for the 8 standard GLUE (\cite{wang2019glue}) subtasks (COLA, MPRC, SST2, RTE, STSB, QQP, MNLI, and QNLI; WNLI is excluded). For each GLUE subtask we run experiments with and without range learning and with per-tensor or per-channel quantization. For all models and exponent bitwidths, per-channel quantization with range learning performed best. For each combination of subtask and exponent bitwidth we use the weight and activation range estimation methods that gave best results in PTQ. We search over a limited range of learning rates, which are subtask dependent, and report results for the best learning rate. We use the learning rates reported in \cite{bondarenko} and optionally extend these to be bigger and/or smaller depending on initial results. In all experiments we use the AdamW optimizer for both model parameters and the quantization range parameters. For the quantization range parameters, we search over learning rates that are 1e1, 1e0 and 1e-1 times as big as the model parameter learning rate. Batch size and numbers of epochs are task dependent and taken from \cite{bondarenko}. We use the per-subtask fine-tuned checkpoints as starting points for QAT. We follow standard fine-tuning practices from \cite{bert} and \cite{huggingface}


\subsubsection{DeepLabV3}
We run experiments using DeepLabV3 \cite{deeplabv3} with a MobileNetV2 backbone. We start from a pre-trained FP32 model. We initialize the FP8 models with the best range estimation method from the PTQ experiments. We use the SBD dataset for improved accuracy, which is a standard option for this model. We use SGD for both the model parameters and the quantization ranges. For each exponent bitwidth we search over learning ranges [1e-1, 1e-2, 1e-3] for the model parameters and [1e-2, 1e-3, 1e-4, 1e-5] for quantization ranges. We train for 200 epochs with a batch size of 8.

\subsubsection{HRNet}
For HRNet (\cite{HRNet}), we largely follow the experimental setup as described by the authors and used in their online repo. We lower the batch size. Since we start from a pre-trained FP32 model and our batch size is lower, we run experiments with lower learning rates. We report results using the best learning rate. We don't learn ranges to reduce memory requirements. We use per-tensor quantization. Best results were obtained by training for 100 epochs with an SGD optimizer and a learning rate of 1e-5 for a batch size of 4, using (all)minmax range estimation for both weights and activations. 

\subsubsection{SalsaNext}
For SalsaNext (\cite{cortinhal2020salsanext}) we largely follow the experimental setup as described by the authors and used in their online repo. We lower the batch size. Since we start from a pre-trained FP32 model and our batch size is lower, we run experiments with lower learning rates. We report results using the best learning rate. We don't learn ranges to reduce memory requirements. We use per-channel quantization. Best results were obtained by training for 75 epochs with an SGD optimizer and a learning rate of 5e-6 for a batch size of 4, using MSE range estimation for both weights and activations. We finetune from a pre-trained checkpoint from \cite{salsanextrepo}. We train and test on the Semantic-Kitti dataset (\cite{semantickitti}). We report mIoU on the Semantic-KITTI validation set.

\section{Experimental Setup Conversion} \label{app:experimental_setup_conversion}

Our goal is to investigate whether models trained in \Efour can be quantized readily to INT8. To ensure a fair comparison, we trained MobileNetV2 and ResNet18 from scratch. This was done to ensure that any differences in weight or activation distributions that may arise from training in FP8 are correctly captured. However, since we saw no difference in the resulting distributions of weight and activation values between models trained using FP8 QAT from scratch and models finetuned using FP8 QAT from a pre-trained FP32 model, we only focused on FP8 QAT finetuned models in our GLUE experiments.

\subsection{ImageNet models}
For ResNet18 and MobileNetV2 QAT we use the experimental setup as described in \cite{graphcore}. For MobileNetV2 we find that we cannot reproduce the exact accuracy number reported in this paper. To alleviate this we expand our search space with various learning rates, include per-channel weight quantization and various methods of range estimation for weights and activations (fixed based on FP8 bitwidth; estimated before training using heuristics as described in \cite{whitepaper}; estimated using heuristics during training; learned as described in \cite{fp8paper}), with minimal results on the resulting accuracy. 

We then quantize the resulting FP8 models to INT8 using per-channel quantization and estimate ranges using MSE range estimation.

\subsection{BERT on GLUE benchmark}
Since we noticed no difference in weight or activation distributions between FP8 fine-tuning of pre-trained FP32 models and FP8 QAT from scratch, we use the fine-tuned FP8 GLUE models form the previous appendix section. We then quantize the learned FP8 weights to INT8 using per-tensor quantization and use MSE range estimation. 

As explained in the main body we find that range estimation methods have a large effect on resulting accuracy. In Table~\ref{tab:range_setting} below we perform an ablation on different range estimation methods for BERT-base on GLUE, and find that MSE range estimation for both weights and activations outperforms other estimation methods by a wide margin on the macro average, and performs best for all individual subtasks except for MRPC, where it is slightly outperformed by minmax range estimation on weights.

\begin{table}[]
\scalebox{0.8}{
    \centering
    \begin{tabular}{| l l |r r r r r r r r | r |}
    \hline
    & & Cola & MNLI & MRPC & QNLI & RTE & SST2 & STSB & QQP & Macro avg \\
    FP8 QAT results & & 62.33 & 84.58 & 87.69 & 91.58 & 72.20 & 93.12 & 89.58 & 89.76 & 83.85 \\ \hline
    Weights & Activations & & & & & & & & & \\ \hline
    current minmax & allminmax & 3.69 & 36.85 & 67.21 & 49.48 & 46.93 & 49.08 & 13.87 & 53.13 & 40.03 \\
    current minmax & MSE & 48.56 & 42.15 & 84.88 & 49.46 & 47.65 & 90.14 & 68.92 & 68.65 & 62.55 \\
    MSE & allminmax & 43.56 & 62.98 & 80.49 & 51.20 & 51.99 & 88.53 & 78.17 & 58.73 & 64.46 \\
    MSE & MSE & 54.72 & 79.93 & 83.84 & 87.57 & 62.09 & 91.17 & 87.34 & 88.34 & 79.38 \\
    \hline
    \end{tabular}
    }
    \caption{}
    \label{tab:range_setting}
\end{table}

\section{INT Quantization Results}  \label{app: aimet_results}

\subsection{INT8 Results}

The Table~\ref{tab:int8results} below compares accuracies of models quantized in INT8 (W8A8: 8-bit weights and activations) and/or W8A16 (8-bit weights, 16-bit activations) using AIMET  (\url{https://github.com/quic/aimet}, \cite{aimet}) vs. FP32 model accuracies for wide variety of networks. Many of these networks can be downloaded from the AIMET model zoo (\url{https://github.com/quic/aimet-model-zoo}).

\begin{table}[]
    \centering
    \begin{tabular}{| l | l r r r |}
    \hline
    Task &	Model &	FP32 KPI & INT8 KPI & Method \\
    \hline
    \multirow{21}{*}{Classification (Top-1\% accuracy)}  & Mobilenetv2 & 71.67\% & 71.14\%	& PTQ \\
	&Resnet18 &	69.75\% &	69.54\%	 & PTQ \\
	&Resnet50 &	76.14\%	 & 75.81\% &	PTQ \\
	&Regnext	& 78.36\%	& 78.10\% &	PTQ \\
	&EfficientNet-lite0	& 75.40\% &	75.36\%	 & PTQ \\
	&Resnet-101	& 77.36\%	& 76.73\%	& PTQ \\
	&DLA-102	& 78.02\% &	77.92\%	& PTQ \\
	&DLA-102x &	78.51\%	& 77.48\%	& PTQ \\
	&SqueezeNet	& 58.09\%	& 57.66\%	& PTQ \\
	&Resnext50	& 77.60\%	& 77.40\%	& PTQ \\
	&Resnex101	& 77.62\%	& 77.40\%	& PTQ \\
	&MnasNet	& 73.40\%	& 73.28\%	& PTQ \\
	&DenseNet	& 74.65\%	& 74.61\%	& QAT \\
	&HRNet-W32	& 78.44\%	& 78.19\%	& QAT \\
	&EFficientNet-B0	& 77.64\%	& 76.68\% & QAT, INT8+W8A16 \\
	&EfficientNetv2-s &	82.90\%	& 82.20\% &	PTQ \\
	&EfficientNet-B4	& 82.67\%	& 81.7\%	& QAT \\
	&Mobilenet-v3	& 74.04\%	& 73.83\%	& PTQ, INT8+W8A16 \\
	&GPU-Net-0	& 78.86\%	& 77.87\%	& PTQ, INT8+W8A16 \\
	&NASNet	& 83.40\%	& 82.30\%	& PTQ \\
	&ConvNext	& 82.37\%	& 81.41\%	& PTQ, INT8+W8A16 \\
        \hline
\multirow{7}{*}{Object Detection (mAP)	} & Mnv2-ssd-lite &	0.687 &	0.686	& QAT \\
	& Centernet (Res50)	& 0.218	& 0.215	& PTQ,INT8+W8A16 \\
	& Yolov3	& 0.653 &	0.649	& PTQ,INT8+W8A16 \\
	& Yolo-v5	& 0.557	& 0.559	& PTQ,INT8+W8A16 \\
	& YoloR	& 0.700	& 0.699	& PTQ \\
	& MobileDet-EdgeTPU	& 0.281 & 	0.279	& PTQ \\
     \hline
\multirow{10}{*}{Segmentation  (mIoU)} & Deeplabv3+	& 78.600 &	78.700 &	PTQ \\
	& Deeplabv3-Xception &	79.700 &	79.000 &	PTQ \\
	& NnuNet	& 87.4/85.9 &	86.4/84.8 &	PTQ \\
	& FFNet	& 81.30	& 80.70	& PTQ \\
	& HRNet-W48	& 81.04	& 80.65	& PTQ \\
	& OCRNet-InverseForm	& 86.31 &	86.21	& PTQ \\
	& HRNet – Inverseform	& 77.81	& 77.00	& PTQ \\
	& PSPNet101	& 81.84	& 81.28	& PTQ \\
	& PSA	& 79.75	& 78.83	& PTQ \\
	& MaskRCNN	& 22.3	& 22.1	& PTQ \\
     \hline
 \multirow{3}{*}{Image SuperResolution  (PSNR (dB))} & XLSR – M7 &	32.66 &	32.58 &	QAT \\
	& SESR – M11	& 32.73	& 32.59	& QAT \\
	& QuickSRNet (L)	& 33.24	& 33.17	& QAT \\
 \hline
\multirow{2}{*}{Pose Estimation  (mAP)} & HRNet-W32 &	0.756	& 0.756 &	PTQ \\
	& HRNet-Posenet	& 0.765	& 0.763	& PTQ\\
 \hline
 \multirow{1}{*}{Action Recognition(Top-1\% accuracy)}  & mmaction2 &	0.785	& 0.778	& PTQ \\
 \hline
  \multirow{2}{*}{LiDAR – Segmentation (mIoU)} &
	SalsaNext	& 55.8	& 55.4	& QAT,INT8+W8A16 \\
	& RangeNet++	& 47.2	& 46.9	& QAT\\
 \hline
\multirow{2}{*}{LiDAR – Object Detection (mAP)} &
	VoxelNet &	0.539	& 0.53	& PTQ,INT8+W8A16 \\
	& PointPillar	& 0.407	& 0.397	 & PTQ,INT8+W8A16 \\
 \hline
 \multirow{1}{*}{Multi-task Network (mIoU/mAP)} &
	YoloP	& 0.89/0.61	& 0.88/0.61	& PTQ,INT8+W8A16 \\
  \hline
  \multirow{5}{*}{Transformers – NLP (GLUE)} &
	BERT &	83.11	& 82.44 &	QAT \\
	& MobileBERT &	81.24 &	81.17 &	QAT\\
	& MiniLM	& 82.23	& 82.63	& QAT \\
	& Roberta	& 85.11	& 84.26	& QAT \\
	& DistilBERT	& 80.71 & 80.26	& QAT\\
  \hline
    \multirow{4}{*}{Transformers – Vision tasks	VIT} &
	VIT	& 81.32\%	& 81.57\%	& QAT \\
	& MobileVIT	& 78.46\%	& 77.59\% &	QAT \\
	& LeVIT &	81.59\%	& 80.84\% &	PTQ \\
	& SWINv2	& 81.81\%	& 81.35\%	& PTQ, INT8+W8A16 \\
    \hline
    \end{tabular}
    \caption{INT8 quantization results, obtained with the AIMET quantization framework. For each network, we picked a setting and method that got us sufficiently close to the original accuracy for commercial purposes. Many models also released in the accompanying model zoo. PTQ indicates network was quantized using PTQ; Unless noted the network is fully quantized in INT8 (W8A8). Some networks are quantized using mixed precision – utilizing mostly INT8 and W8A16 for some small parts of the network.}
    \label{tab:int8results}
\end{table}

\subsection{INT4-W4A8 results}

While INT8 and INT16 are commonly used today, INT4 has been introduced recently to make inference even more energy-efficient. Table~\ref{tab:int4results} below showcases results of quantizing models into W4A8 (i.e., 4-bit weights and 8-bit activations). Most of these models were quantized using QAT. 
			
\begin{table}[]
    \centering
    \begin{tabular}{| l | l r r |}
    \hline
    Task &	Model &	FP32 KPI & INT4 KPI \\
    \hline
    \multirow{4}{*}{Classification (Top-1\% accuracy)}  & 
    ResNet50	& 76.10\%	& 75.63\% \\
    & ResNet18	& 69.75\%	& 69.10\% \\
    & EfficientNet-Lite	& 75.40\%	& 74.44\% \\
    & RegNext	& 78.30\%	& 77.70\% \\
    \hline
        \multirow{4}{*}{Segmentation (mIoU)}  & 
	DeeplabV3 (MNV2) & 	0.729 & 	0.721 \\
	& DeepLabv3 (RN-50)	& 0.7607	& 0.7591 \\
	& HRNet W48	&  0.8104	&  0.8005 \\
	& RangeNet++	&  0.472	&  0.463 \\
 \hline
\multirow{1}{*}{Super-resolution (PSNR (dB))}  & 
         ABPN	& 31.97	& 31.69 \\
        \hline
\multirow{1}{*}{Pose detection}  & 
	 PoseNet (HRNet-32)	& 0.765	& 0.763 \\
\hline
\multirow{1}{*}{Transformers – NLP (GLUE)} &
	BERT	& 83.11	& 82.78 \\
     \hline
    \end{tabular}
    \caption{INT4 quantization results, obtained with the AIMET quantization framework.}
    \label{tab:int4results}

\end{table}

\cleardoublepage
\end{document}